\documentclass{article} % For LaTeX2e
\usepackage{iclr2023_conference,times}

\usepackage{amsmath,amsfonts,bm}

\def\eqref#1{equation~\ref{#1}}
\def\1{\bm{1}}

\DeclareMathAlphabet{\mathsfit}{\encodingdefault}{\sfdefault}{m}{sl}
\SetMathAlphabet{\mathsfit}{bold}{\encodingdefault}{\sfdefault}{bx}{n}

\usepackage[colorinlistoftodos]{todonotes}
\usepackage{enumitem}
\usepackage{amssymb}
\usepackage{hyperref}
\hypersetup{
colorlinks=true,
linkcolor=red,
citecolor=green
}
\usepackage{url}
\usepackage{subfiles}
\usepackage{graphicx}
\usepackage{term_to_use}
\usepackage{boldline}
\usepackage{multirow}
\usepackage{caption}
\usepackage{gensymb}
\usepackage{etoc}

\title{\approach{}: Temporal Alignment Representation with Contrastive Learning}

\author{Yuncong Yang$^{\dagger}$ \quad Jiawei Ma$^{\dagger}$ \quad Shiyuan Huang \quad Long Chen \quad Xudong Lin \quad \\ \textbf{Guangxing Han  \quad  Shih-Fu Chang} \\
Columbia University, New York, NY 10027, USA \\
{\tt\small \{yy3035,jiawei.m,shiyuan.h,cl3695,xudong.lin,gh2561,sc250\}@columbia.edu} \\
}

\iclrfinalcopy
\begin{document}

\maketitle
\let\thefootnote\relax\footnotetext{$^\dagger$Equal contribution. Code Link: \url{https://github.com/yyuncong/TempCLR} }

\begin{abstract}

% Sequence Modelling, Temporal Alignment, Highlight why our method works

Video representation learning has been successful in video-text pre-training for zero-shot transfer, where each sentence is trained to be close to the paired video clips in a common feature space. For long videos, given a paragraph of description where the sentences describe different segments of the video, by matching all sentence-clip pairs,  the paragraph and the full video are aligned implicitly. However, such unit-level comparison may ignore global temporal context, which inevitably limits the generalization ability. In this paper, we propose a contrastive learning framework TempCLR to compare the full video and the paragraph explicitly. As the video/paragraph is formulated as a sequence of clips/sentences, under the constraint of their temporal order, we use dynamic time warping to compute the minimum cumulative cost over sentence-clip pairs as the sequence-level distance. To explore the temporal dynamics, we break the consistency of temporal succession by shuffling video clips w.r.t. temporal granularity. Then, we obtain the representations for clips/sentences, which perceive the temporal information and thus facilitate the sequence alignment. In addition to pre-training on the video and paragraph, our approach can also generalize on the matching between video instances. We evaluate our approach on video retrieval, action step localization, and few-shot action recognition, and achieve consistent performance gain over all three tasks. Detailed ablation studies are provided to justify the approach design.

\end{abstract}

% we first calculate pair-wise sentence-clip matching cost. Under the constraint of temporal order, we then use dynamic time warping to calculate the minimum cumulative cost as the sequence-level distance between the video and paragraph. 

\section{Introduction}

Representation learning on videos has achieved success~\citep{goroshin2015unsupervised,feichtenhofer2021large} in detecting actions in short periods. Recent work has extended it on video-text data~\citep{miech2019howto100m,radford2021learning} to learn a common feature space for zero-shot transfer. In particular, given a paragraph of description, the understanding of \longvideo{s} is increasingly important and may facilitate AI-assistant applications~\citep{grauman2022ego4d,lin2022learning,chen2022weakly}.

A \longvideo{} is usually formulated as a sequence of \shortclip{s}. Given a paragraph, each sentence is used to describe, \ie, paired with, the consecutive video clips in a video segment. 
By matching all sentence-clip pairs~\citep{miech2020end}, the full video and the paragraph can be aligned implicitly.
However, maximizing the agreement between clips and sentences individually (\emph{unit-level}) ignores the context of temporal dynamics, which limits the generalization~\citep{goyal2017something}. 
After all, within one video segment, as the action/event progress at each clip varies, the similarity between the clips and the sentence can be naturally different. 
As such, strictly aligning the sentence with all paired clips, serving as the hard-label, may not always result in an optimal solution. 
% The model may even be trained to focus on background or other spatial information~\citep{buch2022revisiting} which is similar across clip.

\begin{figure}
    \centering
    \includegraphics[width=0.9\textwidth]{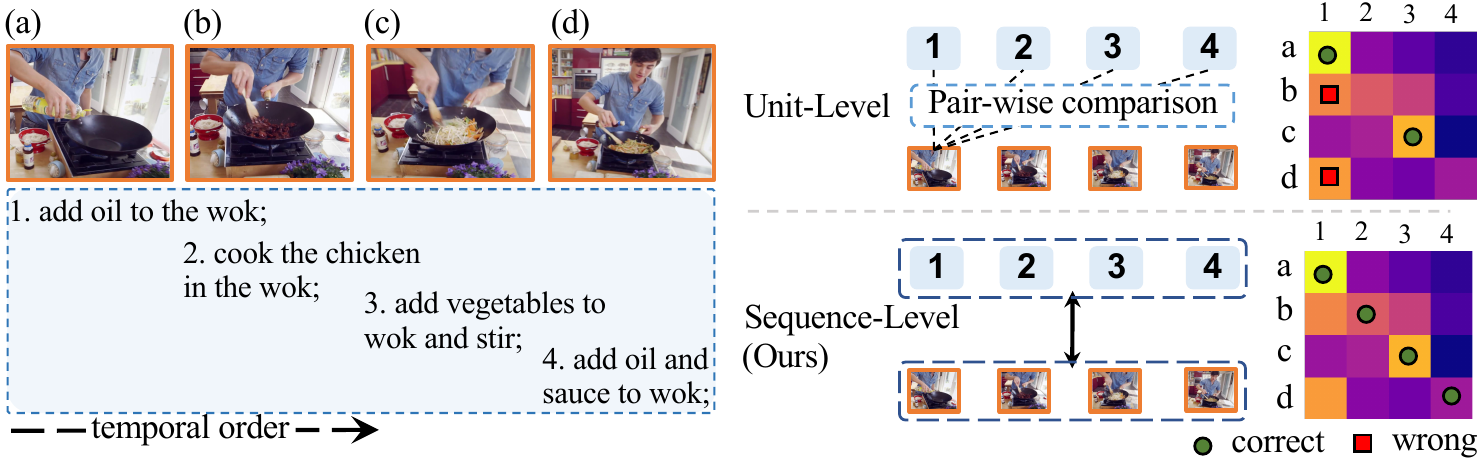}
    \caption{(Left) Given a video and a paired paragraph where the sentences can describe the content in different parts, the temporal orders within the video and the paragraph are consistent. (Right) Conventional methods perform unit-level comparison between sentences and clips pair-wisely and mismatch may occur due to visual similarity. Instead, we directly compare the sequences by considering temporal order such that the temporal succession can be used to align clips and captions.}
    \label{fig:concept}
\end{figure}

To incorporate the temporal correlation across clips, \citet{xu2021videoclip} propose to first fuse the representations over a short period for sentences and video clips separately and then align the fused representations. 
However, such methods only incorporate the \textit{local} temporal information but still does not model the \textit{global} temporal correlation.
As a paragraph is essentially a sequence of sentences, as shown in Fig.~\ref{fig:concept}, the whole \longvideo{} and the paragraph should be explicitly compared and aligned (\emph{sequence-level}).
For a video consisting of multiple steps, \eg, instructional video, the temporal dependence between two distant video clips still exists.
% and the temporal order can thus be utilized. 
% . may exist each step may still depends on the previous steps and the temporal order across the steps can be utilized.
In this way, for a challenging case where two clips are visually similar but are from different segments (clips $\{a,b,d\}$ in Fig.~\ref{fig:concept}), the global context of order can be utilized to avoid the potential mismatching in unit-level matching.

In this work, we study video-paragraph pre-training and propose a framework \approach{} based on sequence-level comparison to explore temporal dynamics.
We directly calculate the distance between full video and paragraph.
Without loss of generality, for the paragraph (anchor) and its paired video (positive sample), the sequence-level distance is the minimum cumulative matching cost over the sentences and clips under the constraint of temporal order and is obtained via dynamic time warping (DTW)~\citep{muller2007dynamic}.
Then we emphasize the unit order which is naturally exhibited within a sequence, and consider the cases where the temporal consistency between video and paragraph is not met.
As a sentence is paired with a segment consisting of multiple clips, we design a negative sampling strategy based on \negkey{}, which shuffles the clips at both unit level and segment level in the paired video. 
% For each anchor, 
% we use the paired video whose clips are randomly shuffled as negative samples and use DTW to find the minimum distance to be maximized.
% In this way, the DTW the corresponding sentence-clip pairs are the most confusing . 
Finally, we apply contrastive learning to maximally align paired video and paragraph.
In this way, we can learn representations for clips and sentences which can perceive global temporal context. Then, from the optimal matching with minimum sequence-level distance, we can pair clips with sentences without being confused by visual similarity.

% assumption the number of the features can be better aligned when the distance is metric

In addition to video-paragraph pre-training, our \approach{} can also be generalized on few-shot action recognition (\votype{}) where each video is classified through the nearest neighbor search according to the sequence distance. 
% As the duration of each action is short, we formulate each video as a sequence of frames. 
% As the frame matching between videos are not annotated, we also replace DTW with OTAM due to a strong assumption in DTW~\citet{cao2020few} to calculate the distance instead.
In summary, the contributions are:
\begin{itemize}[leftmargin=*]
    \item We propose a contrastive learning framework \approach{} to explore temporal dynamics where the learned representation for clips and sentences can facilitate the alignment between sequences.  
    \item Given an anchor, we design a negative sampling strategy based on \negkey{} and shuffle the units in the positive sequence at both segment-level and unit-level. Notably, our method can be generalized to learn representation for both \vltype{} data and \votype{} data.
    \item We conduct extensive experiments on three tasks (\ie, video retrieval, action step localization, and few-shot action recognition) and achieve consistent performance gain to demonstrate the effect of our training strategy. Detailed ablation studies are provided to justify the approach design.
\end{itemize}

\section{Related Work}

\bitem{Contrastive learning} (CT) has achieved success on images~\citep{chen2020improved,he2020momentum} and can cluster samples in the feature space properly.
The main idea is to group different views of the same image/video instance by minimizing the InfoNCE loss~\citep{oord2018representation} while \citet{wang2020understanding} explains it from the aspect of uniformity. Besides, it can be applied on imbalanced dataset~\citep{caron2020unsupervised} and different feature spaces~\citep{grill2020bootstrap,ma2021partner}.

\bitem{Self-supervision on Video} has been studied to learn a good feature extractor. 
Following the idea of temporal jigsaw~\cite{huo2020self}, the order of frames~\citep{misra2016shuffle,lee2017unsupervised} and clips~\citep{xu2019self} can be used for supervision. However, the length of tuple is fixed and too small (\eg, 4) to model temporal context over the full video and may ignore semantic information.
Besides, contrastive learning is also applied for video pre-training where the positive samples can be built by randomly selecting shot video clips~\citep{feichtenhofer2021large}, adjusting the time span~\citep{recasens2021broaden}, and performing spatial-temporal augmentation~\citep{jenni2020video}. In addition, generative approaches~\cite{han2020memory,vondrick2016generating} has also been studied.

% also studied on video representation learning to explore the temporal modelling.
% \citet{feichtenhofer2021large} assumes the visual content of all clips from the same video are consistent when the length is short. Then, \citep{recasens2021broaden} generates views by using different temporal length. Meanwhile, \citet{jenni2020video} comprehensively studies the effect of different temporal augmentation methods for contrastive learning.

\bitem{Multi-modal pre-training for zero-shot transfer} has been studied to connect vision and language. CLIP~\citep{radford2021learning} applies contrastive learning on image-caption pairs.
\citet{yang2022unified} and \citet{li2022grounded} further modify the negative sampling strategy such that the embeddings can be more discriminative at instance-level. 
Then the pre-training is extended to video understanding~\citet{miech2020end,ging2020coot,gabeur2020multi,alayrac2020self,wang2022object}. To improve the performance,  multi-task training has been studied~\citep{li2020hero,luo2020univl}. 
As the description of video content can be noisy, TAN~\citep{han2022temporal} proposes a co-training strategy and use mutual agreement for annotation filtering and VideoCLIP~\citep{xu2021videoclip} proposes a sampling strategy to mitigate the impact of noise in long videos labeling.
Besides video and text, audio is also used to benefit the zero-shot tasks~\citep{chen2021multimodal,shvetsova2022everything} by learning a common feature space.
As an alternative, the Attention mechanism can also be used to fuse the multi-modal information at each layer~\citep{sun2019videobert,su2019vl,chen2020uniter}.

\bitem{Sequence alignment} For each unit in a sequence, along the temporal order, the indices of matched units from aligned sequence pair shall monotonically increase, and averaged distance over matched units is minimized.
Dynamic time wrapping~\citep{muller2007dynamic} is first proposed and canonical time warping~\citep{zhou2009canonical} is then used to align sequences with different feature dimensionality and is applied in deep learning~\citep{sargin2007audiovisual}.
Meanwhile, a pyramid deep architecture~\citep{wang2020alignnet} or attention mechanism~\citep{bishay2019tarn,zhang2020few,ma2019cdsa} can be designed to integrate multi-scale temporal information into a single feature vector.
Besides, regularized by the sequence alignment, pre-training strategies are designed for visual-audio/rhythms synchronization~\citep{cheng2020look,yu2022self}, and video-text alignment~\citep{xu2021videoclip}. 

\section{Approach}

We first provide notation and task formulation in Sec.~\ref{sec:formulation}. Then, we detail the paradigm of our method in Sec.~\ref{sec:approach} and explain how to adapt it for different tasks in ~\ref{sec:application}. 

\subsection{Pre-Training Task Formulation}\label{sec:formulation}

Given an anchor instance $\mathbf{S}_a$ (\ie, a paragraph or a video), we aim to learn a network that can minimize its distance with a positive instance $\mathbf{S}_p$ (\ie, a video paired with the paragraph or a video of the same semantic class).
For each paragraph/video, since its time span can be long, it is typically formulated as a sequence of sentences/video clips. Then, a network is trained to extract a feature for each sentence/video clip, resulting in a sequence of feature embeddings, \ie, $\mathbf{S}_a = \{\mathbf{s}_a^i\}_{i=1}^{N_a}$ and $\mathbf{S}_p = \{\mathbf{s}_p^j\}_{j=1}^{N_p}$, where $\mathbf{s}_a^i, \mathbf{s}_p^j \in \mathcal{R}^d$ are the sequence units, $N_a$ and $N_p$ are the sequence lengths, and $d$ is the dimension of the common feature space. 
In a pair of video and paragraph, the sentence can be mapped with a few consecutive clips, \ie, $\mathbf{s}_a^i$ is matched with $\{\mathbf{s}_p^j\}_{j=t_i^0}^{t_i^1}$ where $\{t_i^0,t_i^1\}$ are the starting and ending feature indexes in $\mathbf{S}_p$ for $\mathbf{s}_a^i$. Then, $N_a$ is not necessarily equal to $N_p$.
In this way, as the intrinsic temporal orders within $\mathbf{S}_a$ and $\mathbf{S}_p$ are consistent, their distance $d_{\{\mathbf{S}_a,\mathbf{S}_p\}}$ should be small. In contrast, two sequences should be distant from each other if they cannot be aligned.

\subsection{Temporal Alignment Representation with Contrastive Learning}\label{sec:approach}

\begin{figure}
    \centering
    \includegraphics[width=0.97\textwidth]{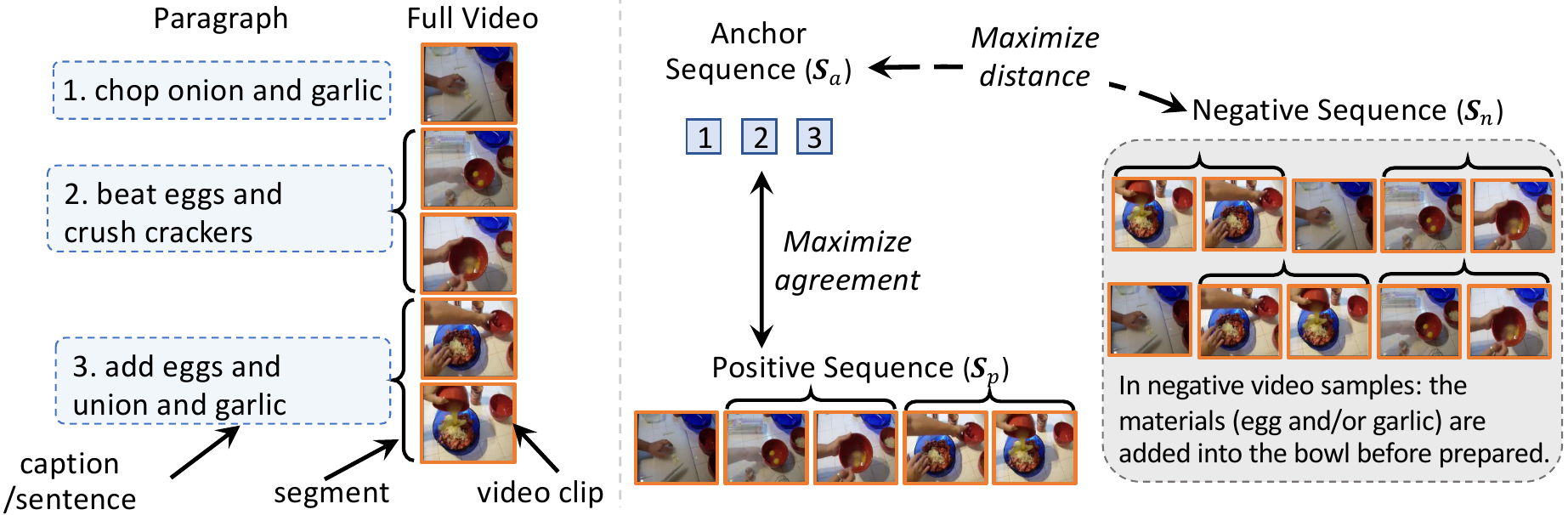}
    \caption{(Left) For each video-paragraph pair, each sentence describes one segment consisting of consecutive video clips. 
    (Right) We set the paragraph (\ie, sequence of sentences) as anchor $\mathbf{S}_a$ and set the paired video (\ie, sequence of clips) as positive sample $\mathbf{S}_p$. Then, we shuffle the units in $\mathbf{S}_p$ to generate a negative sequence $\mathbf{S}_n$ as the temporal succession in $\mathbf{S}_n$ is not consistent with $\mathbf{S}_a$. For clip shuffling, we first alter segments and then optionally shuffle the clips in each segment.}
    \label{fig:approach}
\end{figure}

\revise{In this section, we explain our contrastive(CT)-based training framework \approach{}. We consider the sequences consisting successive steps of a common topic. Then, we propose our negative sampling strategy and choose to use sequence-level comparison to directly calculate the distance.}

The temporal dynamics exhibited in each sequence (\ie, paragraph or video) are representative. To align two sequences, it is also important to ensure the units within the sequences are properly matched. 
For example, as $\mathbf{S}_a$ and $\mathbf{S}_p$ are of temporal consistency, when $\mathbf{S}_a$ and $\mathbf{S}_p$ are aligned, each unit $\mathbf{s}_a^i$ in $\mathbf{S}_a$ should also be matched with $\mathbf{s}_p^j$ in $\mathbf{S}_p$ where $\mathbf{s}_a^i$ and $\mathbf{s}_p^j$ are assumed to be semantically close to each other.
In this way, when two features $(\mathbf{s}_p^i,\mathbf{s}_p^j)$ in $\mathbf{S}_p$ are of two different actions but hard to be distinguished due to visual similarity, the network can still utilize the global temporal order to find a proper unit-level matching between $\mathbf{S}_a$ and $\mathbf{S}_p$ (shown in Fig.~\ref{fig:concept}).

\bitem{Negative Sampling based on \NegKey{}.} In contrast, for a negative sequence $\mathbf{S}_n = \{\mathbf{s}_n^j\}_{j=1}^{N_n}$ which does not preserve temporal consistency with $\mathbf{S}_a$ where $N_n$ is the sequence length, the distance $d_{\{\mathbf{S}_a,\mathbf{S}_n\}}$ between $\mathbf{S}_a$ and $\mathbf{S}_n$ should be high. After all, when $\mathbf{S}_a$ cannot be aligned with $\mathbf{S}_n$, considering the global temporal order, each unit $\mathbf{s}_n^j$ is also distant from any units $\mathbf{s}_a^i$ in $\mathbf{S}_a$ though $\mathbf{s}_a^i$ and $\mathbf{s}_n^j$ can be visually similar (\eg, similar background or foreground objects). 

As such, to mitigate the impact from spatial information, we consider preserving temporal order and aim to obtain representations that facilitates the alignment.
In detail, as shown in Fig.~\ref{fig:approach}, we propose to generate negative sequence $\mathbf{S}_n$ by randomly shuffling $\mathbf{S}_p$ and breaking temporal consistency.
Since the segment-level matching also exists in the paired sequences, we first alter the segment order and then shuffle the units within each segment.
To effectively compare $\mathbf{S}_n$ and $\mathbf{S}_p$ with $\mathbf{S}_a$, we propose a training framework \approach{} based on contrastive learning.

\bitem{Contrastive learning} (\ctloss{}) is a self-supervision approach by learning to group the samples with high correlation. \ctloss{} treats different views of the same instance as correlated (positive) and builds negative pairs by sampling views from different instances. 
As each instance is an image in~\citet{chen2020simple}, each view can be generated by performing augmentation on the low-level pixels, \ie, adding effects such as flipping and color distortion.
In practice, given a set $\mathcal{B}$ with $N_B$ instances, for each instance with one single view $I \in \mathcal{B}$, another view $I'$ is generated and is then used to build a positive pair with $I$ where the view of other instances in $\mathcal{B}$ are used to build negative pairs with $I$. Then, the training objective is to minimize the InfoNCE loss~\citep{oord2018representation},
\begin{equation}\label{eq:Con}
    \conloss{}(I,I',\mathcal{B}_n) = -\log \frac{\exp(\mathbf{z}_{I} \cdot \mathbf{z}_{I'} / \tau )}{\exp(\mathbf{z}_{I} \cdot \mathbf{z}_{I'} / \tau ) + \sum\nolimits_{X \in \mathcal{B}_n} \exp( \mathbf{z}_{I} \cdot \mathbf{z}_{X}  / \tau )}.
\end{equation}
\noindent where $\mathcal{B}_n = \mathcal{B} \setminus \{I\}$, $\mathbf{z}_{X} \in \mathcal{R}^d$ is the feature for instance $X$ after $l_2$-normalization and $\tau$ is a hyperparameter to rescale the affinity scores. 
In this way, \ctloss{} is performing pair-wise comparison where the disagreement between $\mathbf{z}_{I}$ and $\mathbf{z}_{I'}$ is induced by the variation of augmentation.

Note, for our approach \approach{}, $\mathbf{S}_a$ and $\mathbf{S}_p$ served as \emph{two different views} and the \emph{pattern of temporal order that is consistent between $\mathbf{S}_a$ and $\mathbf{S}_p$ is treated as the instance from which the views are augmented}.
Then, we can derive the training objective, \ie, 
\begin{equation}\label{eq:approach}
    \seqloss{}(\mathbf{S}_a,\mathbf{S}_p,\mathcal{S}_n) = - \log \frac{\exp(d_{\{\mathbf{S}_a,\mathbf{S}_p\}} / \tau )}{exp(d_{\{\mathbf{S}_a,\mathbf{S}_p\}} / \tau ) + \sum\nolimits_{\mathbf{S}_n \in \mathcal{S}_n} \exp( d_{\{\mathbf{S}_a,\mathbf{S}_n\}}  / \tau )}
\end{equation}
\noindent where $\mathcal{S}_n = \{\mathbf{S}_{n}^{(i)}\}_{i=1}^{N}$ is the set of $N$ negative sequences derived from $\mathbf{S}_p$.
As a complementary component, the sequences generated from other instances unpaired or uncorrelated with $\mathbf{S}_a$ can also be included in $\mathcal{S}_n$. However, it introduces more computation workload but does not improve performance effectively (analyzed in Sec.~\ref{sec:negative}). As such, for each $\mathbf{S}_a$, we only use the shuffling strategy for negative sample generation. 
By minimizing $\seqloss{}$, $\mathbf{S}_a$ and $\mathbf{S}_p$ are trained to be close to each other while both of them are pushed away from all negative sequences in $\mathcal{S}_n$. Meanwhile, comparing with unit-level comparison between captions and clips, our approach provides a soft supervision on unit-level comparison and we provide detailed comparison in Sec.~\ref{sec:gradient}.

\bitem{Sequence-level distance.} To jointly consider the unit-level matching in sequence alignment, we choose to use Dynamic Time Wrapping (DTW, \citet{muller2007dynamic}), which calculates the minimum cumulative matching costs over units as the sequence distance. The set of matched units is called optimal alignment and is recorded in a matching matrix $M \in \mathcal{R}^{N_a \times N_p}$ with binary indicators.

There are two constraints in DTW, 1) $\forall \mathbf{s}_p^j \in \mathbf{S}_p$ is supposed to be matched with at least one unit in $\mathbf{S}_a$, \ie, $\sum_{1 \leq i \leq N_a} \matchmat{i,j} \geq 1$ for $\forall j \in \{1...N_p\}$, and vice versa; 2) Following the exhibited temporal order, the index of matched units should be monotonously increasing, \ie, if $\mathbf{s}_p^n$ matches $\mathbf{s}_a^m$ where $1 \leq m \leq N_a$ and $1 \leq n < N_p$, $\mathbf{s}_p^{n+1}$ cannot match with any unit in $\{\mathbf{s}_a^i\}_{i=1}^{m-1}$ and $\sum_{1 \leq i < m} \matchmat{i,j} = 0$.
Thus, for $\mathcal{S}_a$ and $\mathcal{S}_p$, when $\matchmat{i,j}=1$, $\mathbf{s}_a^i$ and $\mathbf{s}_p^j$ are matched and assumed to be semantically close to each other. For implementation, we first calculate the pair-wise matching costs $D \in \mathcal{R}^{N_a \times N_p}$ where $D(i,j)$ is the cost, \ie, cosine distance~\citep{singhal2001modern}, between $\mathbf{s}_a^i$ and $\mathbf{s}_p^j$. 
Then, DTW employs dynamic programming and sets a matrix $C \in \mathcal{R}^{N_a \times N_p}$ to record the minimum cumulative cost between $\{\mathbf{s}_a^i\}_{i=1}^{n_a}$ and $\{\mathbf{s}_p^j\}_{j=1}^{n_p}$~\citep{wang2020alignnet,dixit1990optimization}, \ie,
\begin{equation}\label{eq:dtw}
    C(n_a,n_p) = D(n_a,n_p) + \text{min} \{C(n_a-1, n_p-1),~C (n_a-1,n_p),~C(n_a,n_p-1)\}.
\end{equation}
where $1 \leq n_p \leq N_p$ and $1 \leq n_a \leq N_a$. Then, the sequence distance is $d_{\{\mathbf{S}_a,\mathbf{S}_p\}} = C(N_a,N_p)$.
Besides, \citet{cao2020few} propose OTAM, a variant of DTW, to avoid restrict boundary constraint DTW~\citep{muller2007dynamic}. The effect of DTW and OTAM in \approach{} are also studied in Sec.~\ref{sec:experiment} \& \ref{sec:discussion}.

\subsection{Adaptation for Pretraining Tasks}\label{sec:application}

We briefly explain how to apply the paradigm to align the video and paragraph in one semantically-correlated pair (\vltype{}) or videos of the same class (\votype{}) during network training.

\textbf{Video-Paragraph.} For each \longvideo{}, a paired paragraph consisting of short sentences is provided and every sentence, \ie, caption, describes the visual content within a temporal segment. 
Firstly, for paragraph, we learn a text encoder to extract feature $\mathbf{s}_a^{i}$ for each sentence.
Then, for the video, all frames are grouped into non-overlapping consecutive video clips and each clip, serving as a visual token, has $n_f$ frames. Thus, each sentence is paired with multiple consecutive clips.
We use a backbone to extract token embedding for each clip.
% $\Vencoder{}: \mathcal{R}^{h \times w \times 3 \times n_f} \rightarrow{} \mathcal{R}^d$ where $\{h,w\}$ denote the height and width of each frame.
Then, a video encoder is trained to map each clip embedding to the clip feature where all clip features for all sentences are concatenated as $\mathbf{S}_p$.
As the segments for different sentences may have overlap, during training, we will sub-select the sentences to have $\mathbf{S}_a$ such that there is no repeating clips in $\mathbf{S}_p$.

\textbf{Video-only.} Since the duration of action is usually short, we formulate the video as the sequence of frames. As there is no temporal segmentation within each action, we generate the negative sequences by directly shuffling the frames.
In this way, the network is learned in a self-supervised manner.

\section{Experiment}\label{sec:experiment}

We first conduct experiments on video retrieval and action step localization to explain the benefit of \approach{} for zero-shot transfer in long video understanding.
Then, we explain the experiments on few-shot action recognition to indicate the generalization ability from global temporal modelling on new classes.
For the convenience of description, we interchangeably use caption and sentence.
% Since both action step localization and full-video retrieval consider the full video during evaluation. We mainly use these two tasks to examine the importance of utilizing the temporal order/dynamics and show how our approach benefit them.
% the recognition accuracy can be used to indicate the generalization ability of our approach on new classes.

\subsection{Implementation Details of Video-Text Pre-Training}

We follow \citet{xu2021videoclip} and use HowTo100M (HT100M)~\citep{miech2019howto100m} for pre-training. 
As HT100M is too large (~1.2M videos), due to limited computation resource, we build our model on top of the VideoCLIP~\citep{xu2021videoclip} \ie, initialize the VideoCLIP network with its fully-trained model, and randomly select 90k videos (7.5\%) of HT100M to update the network by minimizing $\seqloss{}$. 
% \tocheck{Among the subset, on average, the duration of each video is around 6.5 minutes with about 110 clip-caption pairs. The total text transcriptions is about xxx MB, with 2.4 tokens per second on average.}
% 
VideoCLIP consists of two Transformers~\citep{vaswani2017attention} as encoders for video and paragraph separately. For each clip, they use its S3D feature~\citep{xie2018rethinking} as embedding.
For each sentence, the token embeddings are obtained via a lookup table~\citep{devlin2018bert}.
% \tocheck{for language, one sentence summary}. obtain token embeddings for each sequence via a lookup table~\citep{devlin2018bert} and
% A text encoder $\Tencoder{}: \mathcal{R}^{N_l \times d_w} \rightarrow{} \mathcal{R}^d$ is applied to extract feature for each sentence, where $\{N_l,d_w\}$ are the number of tokens and the dimension of word token respectively.
One MLP layer is set to map clip embeddings and align the dimension of sentence embeddings. 
During pre-training, all encoder parameters are updated.
More experiment details can be found in Appendix.

% and the S3D network is pre-trained on HT100M in a self-supervised manner~\citep{miech2020end}

\subsection{Video-Text Downstream Evaluation}

\subsubsection{Video Retrieval}

\begin{table}[]
    \parbox{.71\textwidth}
    {
        \centering
        \caption{Performance (\%) comparison on full-video retrieval.}
        \resizebox{.71\textwidth}{!}
        {%
            \renewcommand{\arraystretch}{1.13}
            \begin{tabular}{llcrrr}
                \hlineB{3}
                Exp. & (\setup{Background Removed})      & Measure   & R@1       & R@5       & R@10      \\
                \hline
                \revise{1} & MIL-NCE*   & Cap. Avg. & 43.1 & 68.6 & 79.1 \\
                \revise{2} & HT100M*   & Cap. Avg. & 46.6 & 74.3 & 83.7 \\
                \revise{3} & MCN~\citep{chen2021multimodal}       & Cap. Avg. & 53.4 & 75.0 & 81.4 \\
                \hline
                \revise{4} & VideoCLIP$^{\dagger}$ & Cap. Avg. & 74.5 & 94.5 & 97.9 \\
                \revise{5} & VideoCLIP$^{\dagger}$ & DTW          & 56.0 & 89.9 & 96.3 \\
                \revise{6} & \approach{}(Ours)      & Cap. Avg. & 74.5 & 94.6 & 97.0 \\
                \revise{7} & \approach{}(Ours)      & DTW          & \textbf{83.5} & \textbf{97.2} & \textbf{99.3} \\ 
                \hlineB{3}
                & (\setup{Background Kept})      & Measure   & R@1       & R@5       & R@10      \\ \hline
                \revise{8} & VideoCLIP$^{\dagger}$ & DTW & 55.7 & 93.1 & \textbf{98.9} \\
                \revise{9} & \approach{} & DTW & \textbf{70.4} & \textbf{93.8} & 97.9 \\
                \hlineB{3}
                \multicolumn{5}{l}{*:reported in \cite{chen2021multimodal}, $^{\dagger}$: our implementation}
            \end{tabular}
        }\label{tab:full-retrieval}
    }
    \hfill
    \parbox{.275\textwidth}
    {
        \centering
        \caption{Ablation Study on full-video retrieval (\%)}
        \resizebox{.275\textwidth}{!}
        {%
            \renewcommand{\arraystretch}{0.96}
            \begin{tabular}{lc}
                \hlineB{3}
                \multicolumn{2}{l}{\setup{Background Removed}} \\ \hline
                Measure & \approach{} \\
                DTW      & \textbf{83.5}    \\
                OTAM      & 83.4    \\
                \hlineB{3}
                \multicolumn{2}{l}{\setup{Background Kept}}\\ \hline
                Measure & VideoCLIP \\
                DTW      & 55.7    \\
                OTAM      & 53.9    \\ \hline
                Measure & \approach{} \\
                DTW      & 70.4    \\
                OTAM      & \textbf{72.2}    \\
                \hlineB{3}
                \multicolumn{2}{l}{(Metric: R@1 $\times$ 100\%)}
            \end{tabular}
        }\label{tab:metric}
        
        % \approach{}(Ours)      & OTAM          & \underline{83.4} & \textbf{97.3} & \textbf{99.3} \\ 
    }
\end{table}
\begin{table}[]
    \centering
    \caption{Performance (\%) comparison on video retrieval (clip-caption)}
    \resizebox{0.73\linewidth}{!}
    {\renewcommand{\arraystretch}{1.05}
    \begin{tabular}{l|c|rrr}
        \hlineB{3}
        Approach      & backbone   & R@1       & R@5       & R@10      \\
        \hline
        Random                              & -          & 0.0  & 0.2  & 0.3  \\
        MIL-NCE*\citep{miech2020end}        & R152+RX101 & 8.1  & 23.3 & 32.3 \\
        MCN\citep{chen2021multimodal}       & R152+RX101 & 18.1 & 35.5 & 45.2 \\
        MMV\citep{alayrac2020self}          & TSM-50x2   & 11.7 & 33.4 & 45.4 \\
        ActBERT\citep{zhu2020actbert}       & R101+Res3D & 9.6  & 26.7 & 38.0 \\
        MIL-NCE\citep{miech2020end}         & I3D-G      & 11.4 & 30.6 & 42.0 \\
        HT100M\citep{miech2019howto100m}    & S3D-G      & 6.1  & 17.3 & 24.8 \\
        MIL-NCE\citep{miech2020end}         & S3D-G      & 15.1 & 38.0 & 51.2 \\
        MMFT~\citep{shvetsova2022everything} & S3D-G      & \textbf{24.6} & 48.3 & 60.4 \\
        VideoCLIP~\citep{xu2021videoclip}    & S3D-G      & 22.7 & \underline{50.4} & \underline{63.1} \\
        \approach{}(Ours)                   & S3D-G      & \underline{23.3} & \textbf{51.0} & \textbf{64.5} \\              
        \hlineB{3}
    \end{tabular}
    }\label{tab:unit-retrieval}
\end{table}

\bitem{Setup and Metric.} We evaluate our approach under two settings \setup{Full-Video} and \setup{Caption-Clip}. (\setup{Full-Video}) Given a paragraph which contains a set of sentence queries, describing multiple parts of an entire \longvideo{}, the full video should be directly retrieved. To represent the full video, we can either concatenate all clips which have paired captions (\ie, remove background), or directly use the full video with background. 
For retrieval, we can use DTW to measure the distance between the full video and the paragraph directly. Meanwhile, we can still utilize the unit-level comparison, \ie, each caption can be used to match the clip first and the video with the most matched clips will be retrieved, \ie, Cap. Avg.
(\setup{Caption-Clip}) given a sentence description as a query, we retrieve the video clips directly.
For both setups, we use recall as metrics, \ie, R@1, R@5, R@10.

% Given a sentence description as a query, video retrieval aims to find the matching videos from a pool of videos and a widely used setting is to match the sentence with video clips (caption-clip).
% Then, given a set of sentence queries, \ie, paragraph, describing multiple parts of an entire \longvideo{}, \citet{chen2021multimodal} formulates a task to retrieve the full video directly (full-video).

\bitem{Dataset.} We evaluate the model pretrained with our \approach{} on YoucookII\citep{zhou2018towards} without any finetuning (\ie, zero-shot). The evaluation set consists of 3350 caption-clip pairs from 436 videos in total.
The videos existing in YouCookII have been removed from HT100M.
% consists of 2000 cooking videos with a total duration of 176 hours. On average, Each video is about 5.26 minutes and contains \tocheck{XX capations}.

\bitem{Result.} 
(\setup{Full-Video}) 
As summarized in Table~\ref{tab:full-retrieval}, with Cap. Avg as measurement, when background is removed, VideoCLIP has already outperformed MCN clearly (\tablerow{3,4}). However, as VideoCLIP is not trained to explore the global temporal context, the performance drops when DTW is used as measurement.
In contrast, though our approach is only trained with 7.5\% of HT100M full set, benefiting from temporal modelling, \approach{} can effectively improve the performance (\tablerow{5,7}) without hurting the retrieval between clips and captions (\tablerow{4,6}).
Then, we assume no temporal annotation is provided for full-video retrieval and retrieve full video containing background.
% Then, to mimic a more realistic scenario, we do not utilize the temporal annotation of clips and use the paragraph to retrieve the full video containing background.
As the spatial information in background may distract the sentence features, comparing with the scenario when background is removed, the recall by \approach{} drops. 
However, from Table~\ref{tab:full-retrieval} and~\ref{tab:metric}, by using either OTAM or DTW for video-paragraph comparison, our \approach{} can outperform the VideoCLIP baseline consistently.  
(\setup{Caption-Clip}) As summarized in Table~\ref{tab:unit-retrieval}, by minimizing $\seqloss{}$, the attention mechanism in Transformer is trained to embed global temporal information into each clip feature, which may then facilitate the such unit-level retrieval and achieves slight gain over strong baseline VideoCLIP. For MMFT, the extra audio information can be used to benefit the retrieval.

\begin{table}[]
    \caption{Performance comparison (\%) on action step localization for \setup{Zero-shot} (Left) and \setup{Supervised} (right). TFS: training from scratch$^{\dagger}$: Finetune from VideoCLIP with 7.5\% HowTo100M train set.}
    \parbox{.53\textwidth}
    {
        \centering
        \resizebox{.53\textwidth}{!}
        {%
            \begin{tabular}{l|cc}
                \hlineB{3}
                Approach (\setup{Zero-shot}) & TFS & Recall \\
                \hline
                HT100M~\citep{miech2019howto100m} & \checkmark & 33.6 \\
                MIL-NCE~\citep{miech2020end}  & \checkmark  & 40.5 \\
                MCN~\citep{chen2021multimodal} & \checkmark  & 35.1 \\
                DWSA~\citep{shen2021learning} & \checkmark  &  35.3 \\
                UniVL~\citep{luo2020univl}  & \checkmark  &  42.0 \\
                VT-TWINS~\citep{ko2022video} & \checkmark  & 40.7 \\
                VideoCLIP~\citep{xu2021videoclip} & \checkmark  &   33.9 \\
                \hline
                VideoCLIP$^{\dagger}$ & & 33.5 ($\downarrow 0.4$) \\
                \approach{} (Ours)$^{\dagger}$ & & \textbf{36.9} ($\mathbf{\uparrow 3.0}$) \\
                \hlineB{3}
            \end{tabular}
        }
    }
    \hfill
    \parbox{.47\textwidth}
    {
        \centering
        \resizebox{.47\textwidth}{!}
        {%
            \renewcommand{\arraystretch}{1.15}
            \begin{tabular}{l|c}
                \hlineB{3}
                Approach (\setup{Supervised})  & Recall \\ 
                \hline
                Alayrac~\citep{alayrac2016unsupervised}          & 13.3 \\
                Zhukov~\citep{zhukov2019cross}           & 22.4 \\
                Supervised~\citep{zhukov2019cross}       & 31.6 \\
                VideoCLIP~\citep{xu2021videoclip}        & 47.3 \\
                \approach{} (Ours)             & \textbf{52.5} ($\uparrow \mathbf{5.2}$) \\
                \hlineB{3}
                Approach (Few-shot)  & Recall \\ 
                \hline
                VideoCLIP w/  20\%               & 41.1 \\ 
                \approach{} (Ours) w/  20\%             & \textbf{42.8} \\
                \hlineB{3}
            \end{tabular}
        }
    }\label{tab:localization}
\end{table}

\subsubsection{Action Step Localization}

\bitem{Setup and Metric.} Each video is associated with a \underline{T}ask consisting of multiple steps (\ie, sentences). 
Then, each video frame should be assigned with the corresponding step and we use recall as metric. 

\bitem{Dataset.} We perform evaluation on CrossTask~\citep{zhukov2019cross} and the test set contains 1690 annotated videos over 18 \underline{T}asks.
We first apply our model pre-trained on HT100M on CrossTask test set for zero-shot evaluation.
Then, following \citet{xu2021videoclip}, we finetune the VideoCLIP model on 540 videos with our $\mathcal(L)_{seq}$ and then evaluate the finetuned model on test set (\setup{Supervised}).

% consisting of 83 different \underline{T}asks over 4.7K videos. 
% From the official split, the 
% and the train set then consists of 65 \underline{T}asks from ~3K videos and there is no \underline{T}ask overlap between the train set and the test set. The steps candidates during inference are not seen during training either.
% Thus, we can apply our model pre-trained on HT100M on CrossTask test set for zero-shot evaluation.
% Meanwhile, we can finetune the VideoCLIP model on CrossTask train set with our $\mathcal(L)_{seq}$ and then evaluate the finetuned model on test set (\setup{Supervised}).

\bitem{Result.} (\setup{Zero-shot}) As shown in Table~\ref{tab:localization}, when we update VideoCLIP with the 7.5\% subset using its original loss (\tablerow{8}), the model overfits and the performance drops slightly. However, for \approach{}, by adding loss $\seqloss{}$, we can still improve the recall from 33.9 to 36.9.
(\setup{Supervised}) VideoCLIP has shown strong performance, but \approach{} can still increase the recall to 52.5.
Furthermore, by finetuning on only 100 videos (20\% of train set), our approach can effectively improve the performance, which also demonstrates the benefit from exploring the temporal order modelling.

% As shown in Table~\ref{tab:localization}, by adding loss $\seqloss{}$, our method achieves consistent (9\%) performance gain over VideoCLIP baseline in both setups.
% (\setup{Zero-shot}) when we keep updating VideoCLIP with the 7.5\% training data using its original loss, the model may overfit and the performance will drop slightly (\tablerow{7,8}).
% In contrast, for \approach{}, we can still improve the performance from 33.9 to 36.9.

% Though VideoCLIP has shown strong performance after finetuning, our method can still increase the performance from 47.3 to 52.5.  
% Meanwhile, the two values in $R_{2}$ are obtained by using different feature embedings.

\subsection{Few-shot Action Recognition}

\begin{minipage}{\textwidth}
  \begin{minipage}[]{0.47\textwidth}
  \bitem{Setup and  Metric.} Given a \underline{T}ask where each class has only $N_s$ labeled video as reference (\ie, $N_s$-shot), we classify a test video by comparing its average distance with labeled videos of each class through nearest neighbor search.
  Following the protocol \citep{zhu2018compound}, we first pre-train the model on a dataset of classes $\mathcal{C}_{b}$ and directly evaluated on few-shot tasks which are sampled from another dataset of classes $\mathcal{C}_{n}$ and $\mathcal{C}_{b} \cap \mathcal{C}_{n} = \O$.
  \end{minipage}
  \hfill
  \begin{minipage}[]{0.5\textwidth}
    \centering
    \captionof{table}{Performance (\%) on action recognition.}
    \resizebox{0.95\textwidth}{!}
    {
        \begin{tabular}{l|cc}
            \hlineB{3}
            Approach & 1-shot & 5-shot \\
            \hline
            TSN++*   & 33.6 & 43   \\
            CMN++*   & 34.4 & 43.8 \\
            RTRN++*  & 38.6 & 48.9 \\
            OTAM~\citep{cao2020few}    & 42.8 & 52.3 \\
            TRX~\citep{perrett2021temporal}     & 42.0   & 64.6 \\
            MTFAN~\citep{wu2022motion}   & {45.4} & 60.4 \\
            \approach{} (Ours) & \textbf{47.8} & \textbf{67.7} \\
            \hlineB{3}
            \multicolumn{3}{l}{*:Results are reported in \citet{cao2020few}}
        \end{tabular}\label{tab:fsar}
    }
  \end{minipage}
\end{minipage}
\bitem{Dataset}. We use sth-sth V2 \citep{goyal2017something} for experiment and follow the subset split in \citet{cao2020few}.
The subset contains 100 classes where $|\mathcal{C}_{b}| = 64$ and $|\mathcal{C}_{n}|$ is 24 (12) classes are for evaluation (validation).
During evaluation, each \underline{T}ask contains 5 classes and has 15 test videos per class while $N_s = \{1,5\}$.
Finally, we report the mean accuracy over 10k \underline{T}asks. As only using spatial content is unreliable, temporal modelling is thus specifically required for correct prediction.

\bitem{Results} For fair comparison, we first use a ResNet50 model pretrained on ImageNet~\citep{deng2009imagenet} to extract embedding for each frame which are fed into a Transformer to obtain features in $\mathbf{S}_a$. As for $\mathbf{S}_p$, we set a linear layer to process each frame embedding. Then, we use the pre-trained model as initialization and follow the training of OTAM and TRX. More details can be found in Sec.~\ref{sec:downstream-supp}. \approach{} differs from \citet{cao2020few} by applying self-supervision and using a shuffled version of $\mathbf{S}_p$ as $\mathbf{S}_n$, while they apply meta-training~\citep{snell2017prototypical} between video instances. TRX also employs a Transformer but only learns the temporal order between a pair/triple of frames and the learning efficiency is limited. In contrast, TempCLR directly models the temporal context.

% As the classes on sth-sth V2 typically requires the modelling of temporal dynamics while purely rely on spatial information cannot work, it further demonstrates our approach is effective in modelling temporal context.
% for ResNet50 baseline, we directly uses the embeddings extracted by the ResNet50 and use DTW as measurement for action recognition.

% For videos, a backbone network is used to extract embedding for each clip, \ie, video token, where the clip embeddings are then fed into the 6-layer video transformer. 
% For each sentence in the paragraph, they first obtain embedding for each token \tocheck{though embedding look-up used in BERT~\citep{devlin2018bert}} and then send the tokens of each sentence into a 12-layer Transformer.
% VideoCLIP uses S3D feature~\citep{xie2018rethinking} as the clip embeddings and the S3D network has been pre-trained on HowTo100M~\citep{miech2019howto100m} in a self-supervised manner~\citep{miech2020end}.
% To align the dimension of embeddings, VideoCLIP additionally set a MLP with the Transformer to map the clip emedding dimension from 512 to 768.
% orthogonal to the temporal module network design
\section{Discussion}\label{sec:discussion}

\subsection{Unit Matching in Sequence Alignment}\label{sec:unit-seq} 
When we directly measure the global distance between video and paragraph, it is also very important to ensure the matched caption-clip pairs are semantically close to each other.
As visualized in Fig.~\ref{fig:visualization-main}, for the full video directly retrieved by a paragraph, our approach can also correctly match all captions with the video clips. 
However, if we only rely on the unit-level similarity, the high visual similarity can cause false prediction. More visualization can be found in appendix.

In addition, as shown in Table~\ref{tab:ablation}(a), we check percentage of correctly matched clip-caption pairs averaged over all videos.
Then, our \approach{} can correctly match more caption-clip pairs than VideoCLIP when the video is compared with paragraph using DTW.
In this way, given a paragraph, our approach can also mitigate the distraction from unpaired videos.

\begin{minipage}{\textwidth}
  \begin{minipage}[b]{0.63\textwidth}
    \centering
    \includegraphics[width=\textwidth]{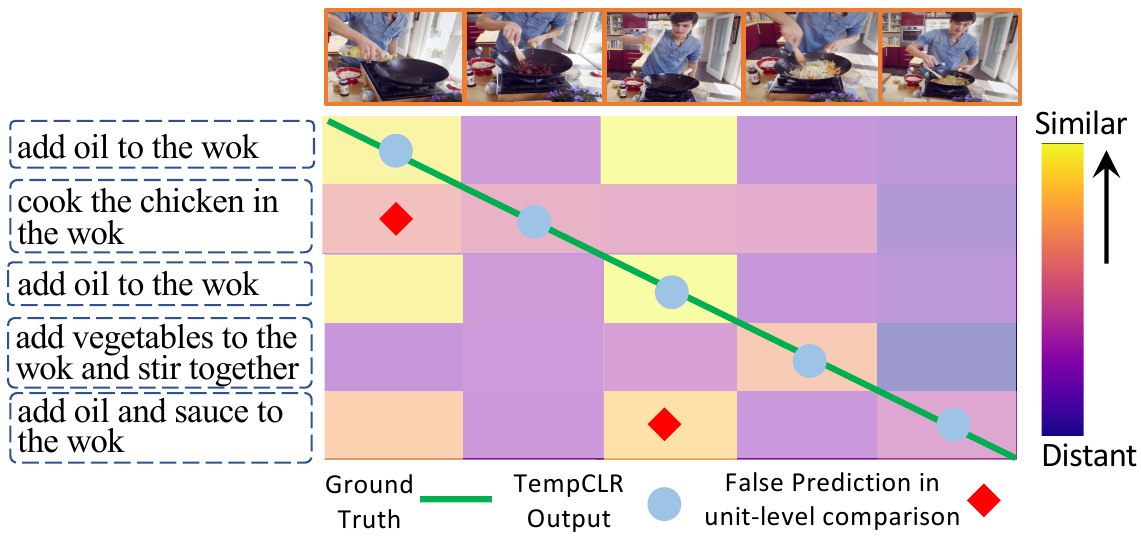}
    \captionof{figure}{Visualization of video retrieval. \approach{} correctly match all clip-caption pairs while retrieving the full video.}
    \label{fig:visualization-main}
  \end{minipage}
  \hfill
  \begin{minipage}[b]{0.37\textwidth}
    \centering
    \renewcommand{\arraystretch}{1}
    \resizebox{\textwidth}{!}
    {
        \begin{tabular}{c|cc}
            \multicolumn{3}{l}{(a) Full-Video Retrieval}\\
            \hlineB{3}
            Approach & \multicolumn{2}{c}{Matched Clip-Caption} \\
            \hline
            VideoClip &  \multicolumn{2}{c}{79.6\%} \\
            \approach{} & \multicolumn{2}{c}{91.6\%} \\
            \hlineB{3}
            \multicolumn{3}{l}{ }\\
            \multicolumn{3}{l}{(b) Ablation study (\%) on}\\
            % \multicolumn{3}{l}{action step localization}\\
            \hlineB{3}
            PT \textbackslash{} DS & w/o $\seqloss{}$ & w/ $\seqloss{}$ \\
            \hline
            VideoClip &  47.3 & 52.5 \\
            \approach{} &  49.3	& 52.3 \\
            \hlineB{3}
            \multicolumn{3}{l}{PT: Pre-train, DS: Downstream}\\
        \end{tabular}
    }
    \captionof{table}{Ablation study.}
    \label{tab:ablation}
    \end{minipage}
  \end{minipage}

% \begin{figure}
%     % \centering
    %  can clearly ahcieve one-to-one mapping after training with temporal order
%     \caption{Visualization of alignment on YouCookII. Our approach can clearly ahcieve one-to-one mapping after training with temporal order. The full result can be found in Appendix}
%     \label{fig:visualization-main}
% \end{figure}

\subsection{Negative Sample Selection}\label{sec:negative}

% Our approach follows the contrastive learning framework and minimizes the InfoNCE \citep{oord2018representation} loss. 
For each anchor sequence $\mathbf{S}_a$, we generate negative samples $\mathbf{S}_n$ from the positive sample $\mathbf{S}_p$ by 1) considering the difference regarding temporal granularity and 2) shuffling the order to break the temporal consistency. Then, we discuss alternative strategies.

\bitem{Video-Paragraph.} For each paragraph (anchor), in addition to shuffle the segments first and then shuffle the clip embeddings within each segment (\textit{seg-unit}), we can also only shuffling the segments while maintaining the clip order within each segment (\textit{seg-only}).
Meanwhile, an intuitive strategy is to use the video unpaired with the anchor to build negative samples (\textit{unpaired}) or combine it with \textit{seg-unit} \textit{joint}.
In addition, we can aggressively shuffle all clips embeddings in $\mathbf{S}_p$ (\textit{all-unit}), or only shuffle the clip embeddings within each segment (\textit{within-seg}).

As shown in Table~\ref{tab:negative}, we compare the performance on CrossTask under \textit{Supervised}. 
(1) Since we use VideoCLIP as initialization and the model has been trained to distinguish clips from different video instances, the sequence-level distance of negative samples in \textit{unpaired} has already been high. Furthermore, as the temporal consistency hardly exists between video and unpaired paragraph, the gain by \textit{unpaired} is inevitably limited.
(2) According to \tablerow{2-4}, since each caption is supposed to align with all of the clips in the paired segment, breaking the order of segments in $\mathbf{S}_p$, \ie, \textit{seg-only} and \textit{seg-unit}, is essential in facilitating model training.
From Sec.~\ref{sec:unit-seq}, \textit{when two sequences are aligned using DTW, the matched units are also semantically close to each other}. Thus, when 
\begin{minipage}{\textwidth}
  \begin{minipage}[]{0.68\textwidth}
  $\mathbf{S}_a$ is compared with $\mathbf{S}_n$, the matched units can indicate the clip features which may cause the most confusion in caption-clip matching due to high visual similarity. In this way, by minimizing $\seqloss{}$, the network is then trained to distinguish those clip features which may hurt the alignment between $\mathbf{S}_p$ and $\mathbf{S}_a$.
  Then, when the segment order is preserved, comparing with VideoCLIP baseline (\ie, 47.3), \textit{within-seg} results in worse generalization as the confusing clips across segments are not detected and the model can be trained to overfit to trivial difference between clips under the same segment. In contrast, when the segment order is broken, shuffling the clip
  \end{minipage}
  \hfill
  \begin{minipage}[]{0.3\textwidth}
    \centering
    \captionof{table}{Ablation study (\%) of negative sampling.}
    \resizebox{\textwidth}{!}
    {
        \renewcommand{\arraystretch}{0.93}
        \begin{tabular}{lc|c}
            \hlineB{3}
            Exp. & Strategy & Recall \\
            \hline
            1 & un-paired & 48.0\\
            2 & within-seg & 46.4 \\
            3 & seg-only & \underline{52.1} \\
            4 & seg-unit & \textbf{52.5} \\
            5 & all-unit & 49.3 \\
            6 & joint & \textbf{52.5} \\
            7 & visual-anchor & 52.0 \\
            \hlineB{3}
        \end{tabular}\label{tab:negative}
    }
  \end{minipage}
\end{minipage}
order within each segment further can serve as data augmentation, which can improve the recall slightly from 52.1 to 52.5. Furthermore, (3) 
\textit{all-unit} only introduce limited gain since the continuity across clips in one segment is broken and it is too hard for the model to learn. Combining \textit{unpaired} and \textit{seg-unit} in \textit{joint} does not provide clear gain over \textit{seg-unit}. However, we think the main reason is that VideoCLIP has been well-trained for instance discrimination and believe \textit{joint} is still necessary when \approach{} is trained from scratch.
Lastly, we can also shuffle the sentence embeddings w.r.t a video (\textit{visual-anchor}) which is equivalent to \textit{seg-only} and achieve reasonably high performance.
% instead of setting the paragraph as anchor, 
% As such, the performance by \textit{visual-anchor} is also high.

\bitem{Video-Only.} As an alternative, we can also shuffle the frame features of other videos as $\mathbf{S}_n$ and keep training the model in a self-supervised manner. 
However, since the distance between different video instances has already been high, the model is not well learned and the accuracy is 38.2. 

% As each video instance is assigned with an action class label but no segment annotation is provided, for each anchor video, besides shuffling all frame embeddings in $\mathbf{S}_p$ (\textit{w/o label}, used in our approach), the alternative negative generation strategy it to select video samples from different categories (\textit{w/ label}).

\subsection{Component Analysis and Ablation Study}\label{sec:ablation}

\bitem{Modelling of global temporal context} has been studied for long-video understanding. A popular way is to employ Transformer architecture to model the correlation between clips automatically.
However, by explicitly modelling the temporal orders, as demonstrated in Sec.~\ref{sec:experiment}, \approach{} is capable to provide consistent gain over three tasks under six different setups from the strong baseline.
Thus, the comparison with VideoCLIP already serves as ablation studies to demonstrate the importance of explicit regularization for temporal modelling.
After all, the Attention mechanism may require many data to fully understanding the temporal correlation. 
In addition, specifically for videos, the labels are noisy such as misalignment between ASR transcription and long video~\citep{miech2020end}, the attention modelling can also be distracted by background which hurts the training efficiency.
As such, our approach provides a general framework aiming to utilize temporal context.
% Similarly, we notice that \citet{huang2021cross} improves localization precision by regularizing the temporal ordering between nearby actions. However, ours approach targets on the modelling of the full video and provides a more general framework, \ie, shuffling all captions (clips) in one paragraph (\longvideo), implicitly mining the confusing cases resulted from the visual similarity by comparing $\mathbf{S}_a$ and $\mathbf{S}_n$.
%  and can also avoid the confusing cases

\bitem{\textit{Supervised} on CrossTask}. As summarized in Table~\ref{tab:ablation}, we use CrossTask study the effect of $\seqloss{}$ in pre-training (PT) stage and downstream finetuning (DS) stage. 
For \approach{}, \ie, with $\seqloss{}$ in PT, as the model has been trained to model global temporal order, finetuning without $\mathcal{L}_{seq}$ can also improve the recall.
Meanwhile, though the temporal pattern learned PT may not exactly the same of data in DS, as finetuning with $\mathcal{L}_{seq}$ in DS is very important for down-stream evaluation, the performance are comparable when either VideoCLIP or \approach{} is used for initialization. 

\bitem{DTW Vs. OTAM}. From Table.~\ref{tab:metric}, as the temporal annotation is given in video-paragraph pretraining, using DTW or OTAM achieves similar performance. However, even when annotation is not given in Video-only task, employing either OTAM or DTW does not impact the performance significantly (47.7 for 1-shot). More details can be found in the appendix.

% video re-localization (ECCV 2018)
% future direction: full-filling the 

% After all, even though the position embedding~\citep{vaswani2017attention,devlin2018bert} is set, the attention mechanism in still treat every token equally by default and may still requires a lot of trianing data to learn the modelling.
% Similarly, we notice \citep{huang2021cross} has consider temporal ordering between nearby action steps and thus proposes regularization for action localization. However, ours generalize to multi-instance under InfoNCE loss form and specifically focus on the hard negative resulted from the visual similarity to better distinguish the spatial and temporal.
% VideoCLIP is optimized by minimizng the InfoNCE~\citep{chen2020simple} between captions and clips, and does not clearly model the temporal orders during trianing.
% As VideoCLIP has already been a strong baseline, we choose it for comparison to better demonstrate the importance of temporal modelling. 

\section{Conclusion}
In this paper, we have proposed \approach{}, a contrastive learning framework to align the temporal dynamics exhibited in video-paragraph pre-training, where the paragraph/video can be represented as a sequence of sentences/clips and are learned to match their global content following the consistent temporal order. Specifically, to encourage temporal modeling over the full sequences and the effectiveness of explicit sequence comparison, we propose to generate negative samples by shuffling clips of the paired video from different temporal granularities. In this way, we can adjustably maximize the agreement between sequences with temporal order consistency and maximize the distance between unaligned sequences with different temporal coherency. Our \approach{} achieves consistent performance gain on three tasks under six different setups, which experimentally demonstrated the effectiveness of our framework. We also provide analysis to validate our design choice of negative sampling, which is shown to both benefits the sequence-level alignment and the unit-level matching.

\noindent{\bf Acknowledgement}
\noindent This material is based on research sponsored by Air Force Research Laboratory (AFRL) under agreement number FA8750-19-1-1000. 
The U.S. Government is authorized to reproduce and distribute reprints for Government purposes notwithstanding any copyright notation therein. 
The views and conclusions contained herein are those of the authors and should not be interpreted as necessarily representing the official policies or endorsements, either expressed or implied, of Air Force Laboratory, DARPA or the U.S. Government.

% We first represent the paragraph (full video) as a sequence of sentences (video clips). 
% Without loss of generality, for each paragraph serving as anchor, we use the paired sequence of clips as a positive sample and then shuffling those clips to generate the negative samples.
% In this way, we use the paragraph and the full video to represent two views of the same temporal order pattern and maximize their distance with the negative sequences as the temporal orders are not consistent.
% Then, according to the ablation study of negative sampling strategy, for video-paragraph data, as each sentence is paired with a segment consisting of a few consecutive clips, it is important to shuffle along segment and aggressively shuffling all clips directly cannot help with the model training.
% Our \approach achieves consistent performance gain on three tasks under six different setups, which experimentally demonstrated the effectiveness of our approach. Qualitative visualization is also provided to show that our approach can both benefit the sequence-level alignment but can also benefit the unit-level matching between caption and clips. 



\bibliography{iclr2023_conference}
\bibliographystyle{iclr2023_conference}

\newpage 
\appendix

\hypersetup{linkcolor=black}
\section{Appendix}

{
\hypersetup{linkcolor=black}
\localtableofcontents
}
\newpage
% \tableofcontents

\subsection{Terminology Explanation}\label{sec:term}
In this paper, we study the video-text pre-pretraining for long-video understanding, and specifically focus on the video-paragraph pre-training. As such, we interchangeably use video-text pre-training and video-paragraph pre-pretraining.
Here, the paragraph is essentially a set of captions and each caption is used to describe one part, \ie, segment, in the full video. As such, according the order of video segments, the captions can be organized using the same order and represented as a form of paragraph.
As the form of caption is a short sentence, we interchangeably use caption and sentence, such as sentence-caption pair and caption-clip pair.

For pre-training on HT100M~\citep{miech2019howto100m}, due to the limitation of computational resources, we essentially finetune from the pre-trained VideoCLIP model using 90k videos (7.5\%) of HT100M. However, for the convenience description, we still use pre-training for the training on HW100M and use ``finetuning'' as fine tuning on the train set of downstream task, \ie, CrossTask (\setup{Supervised}).

In downstream evaluation such as action step localization and few-shot action recognition, we use ``\underline{T}ask'' to indicate the a localization task (matching each caption in a to different parts in the video) and one few-shot task (classify test videos among the few-shot classes of interest), which is different from ``task'' used to indicate each of the three task types.

\subsection{Implementation Details for Training}

\subsubsection{Implementation Details for HW100M Pre-training}

\paragraph{Architecture.}
VideoCLIP~\citep{xu2021videoclip} consists of two Transformer architectures for video and paragraph separately. For videos, a backbone network is used to extract embedding for each clip, \ie, video token, where the clip embeddings are used as the input of the six-layer video transformer. 
For each sentence, VideoCLIP first tokenize the sentences and obtain the token embedding (\eg, the embedding for each word) from embedding lookup~\citep{devlin2018bert}.
For implementation, VideoCLIP use S3D feature~\citep{xie2018rethinking} as the clip embeddings as the S3D network has been pre-trained on HowTo100M~\citep{miech2019howto100m} in a self-supervised manner~\citep{miech2020end}.
As the embedding dimensions for clips (512) and language tokens (768) are different, an MLP is set preprocess the clip embeddings and increase their dimension. As such, the trainable network components of VideoCLIP includes two Transformer and one MLP layer. During our pre-training, we update the all parameters.

\paragraph{Dataloading.}
As the annotation of temporal segments for long videos is noisy and ASR transcription can also be misaligned with long video~\citep{miech2020end}, for each caption, VideoCLIP has designed a strategy 
\begin{itemize}[leftmargin=*]
    \item Extend the timestamps for the caption of interest and the period after extension may include other captions. Then, the newly included captions is concatenated with the caption of interest which are fed in the 12-layer text Transformer.
    \item Average the Transformer outputs for all of the tokens to obtain one representation for the caption.
    \item Use the new timestamps to select clip and feed them into 6-layer video Transformer.
    \item Fuse the information within the segment by averaging the output of video Transformer, \ie, clip features, and obtain a segment representation. 
\end{itemize}

For our approach \approach{}, we reuse its data loading pipeline here to mitigate the potential issue from the noisy annotation. 

\paragraph{Training implementation.} As the features of clips in the same segment has been averaged to obtain one segment representation, during pre-training of our \approach{} on HW100M, we directly shuffle the segment features in $\mathbf{S}_a$ without shuffling the clips features as the segment feature is independent of the clip feature orders. Meanwhile, during the pre-training, we set the batch size of captions is 256 and the captions are sampled from 16 long videos.
As we trained from the VideoCLIP pre-trained model, when all parameters are trained, we set the learning rate as $1e^{-5}$. 
Then, for comparison, as an ablation study, we only update the parameters in the norm layers such as layer norm and the learning rate is set as $2e^{-5}$.
In both case, we train the model on 2 NVIDIA TiTAN RTX GPUs (each with 24 GB memory) for 10 epoches within two hours and use the default hyper-parameters in Adam~\citep{kingma2014adam} optimizer with betas of (0.9, 0.98).

\subsubsection{Implementation Details for CrossTask Finetuning}

\paragraph{Architecture.} Refer to VideoCLIP architecture.

\paragraph{Dataloading.}
When the model is finetuned on CrossTask for action step localization, we still keep the embedding of each clip and do not fuse them. As such, we can apply the shuffling strategy described in Fig.~\ref{fig:approach} and the ablation study is provided in Sec.~\ref{sec:negative}.
\begin{itemize}[leftmargin=*]
    \item For each video, we sample clips by sliding window with window size 16 and each clip takes 32 frames, and load all clips of the video into a batch
    \item Put all the action steps for the given task into the text encoder to get action step representations. Put the sampled clips to the video encoder and expand the clips to full video in frame level (by returning hidden state).
    \item Filter out background frames and only keep frames with action step labels.
    \item Generate negative samples by swapping the action step representations and get the alignment loss. 
\end{itemize}

\paragraph{Training implementation.} The learning rate is set as $5e^{-5}$. We use the default hyper-parameters in Adam~\citep{kingma2014adam} optimizer with betas of (0.9, 0.98)

\subsubsection{Implementation Details for Video Pre-Training}

\paragraph{Architecture.} We use a ResNet50 pretrained on ImageNet to extract embedding for each frame. The Transformer architecture has 6 layers and output the feature for each frame embedding. 
The linear layer that is set to process each frame embedding to obtain $\mathbf{S}_p$ is also jointly learned.

\paragraph{Data Loading.} Then, we follow~\citet{cao2020few} to first determine the starting and ending frame index as well as the sampling rate, then we use the selected frame embeddings as the sequence representing the video.
According to the starting and ending frame index as well as the sampling rate, we can determine the frames to be sampled. 

\paragraph{Training implementation.} During training, we only train the Transformer while the ResNet50 is not trained. However, for the methods compared in the Table.~\ref{tab:fsar}, the ResNet50 are tuned. 
The Transformer is trained from scratch and it uses adam optimizer with default hyper-parameter setting. We use cosine annealing schedule and the initial learning rate is 0.001. 
Meanwhile, we use the weighted sum (0.3, 0.7) of VideoCLIP loss and our $\mathcal{L}_{seq}$ as the final training objective. Since we update the network from the well-trained VideoCLIP parameters, maintaining the original loss of VideoCLIP is to keep the stability of network training. The weight here is also an empirically value that can provide reasonable performance gain and we fix it during the pre-training.
% Meanwhile, we use the weighted sum (0.3, 0.7) of VideoCLIP loss and our $\mathcal{L}_{seq}$ as the final training objective. In this way, we aim to make the model be aware of the temporal order context during training but still maintain the ability to match temporal information fused in a short time span. In this way, the performance gain by our approach.

\subsection{Downstream Task Explanation}\label{sec:downstream-supp}

\subsubsection{Video Retrieval}

Video retrieval has been widely studied and its most popular setting is go retrieve video clips given the sentence/caption as a query, (\ie, caption-clip).
Recently, to formulate a more realistic scenario, \citet{chen2021multimodal} formulated full-video retrieval, \ie, given a set of caption queries, \ie, paragraph, describing multiple parts of an entire  \longvideo{}, the task then aims to retrieve the full video according to the paragraph.
However, for each long video, \citet{chen2021multimodal} directly concatenate all video clips which has paired sentences. Then, a strong assumption is applied that the background can be removed according to the temporal annotation.
As such, we take a step further to consider the full-video retrieval without any temporal annotation and the background are included.
The performance is measured by recall, the number of captions/paragraphs that can correctly retrieve the clips/videos.
For each language query, we will rank the retrieving result according to the similarity and a correct assignment means the video is correctly retrieved among the videos with top k high similarity, \ie, R@K, and we have $k=\{1,5,10\}$

\subsubsection{Action Step Localization}

This task assumes that each video is associated with a \underline{T}ask consisting of multiple steps.
During inference, given a \underline{T}ask and its corresponding step candidates, each video frame/clip is supposed to be assigned to the corresponding step. This assignment is determined according to the similarity between the frame/clip feature with the caption feature, \ie, in a form of classification task. 
Then, we directly use recall R@1, \ie, number of step assignments that fall into the correct ground truth interval, divided by the total number of steps in the video, as the metric.

\subsubsection{Few-shot Action Recognition}

For few-shot learning, we are first given a \textit{base} dataset of classes $\mathcal{C}_{base}$ and a \textit{novel} dataset of classes $\mathcal{C}_{novel}$ where the two class sets are \emph{disjoint}, \ie, $\mathcal{C}_{base} \cap \mathcal{C}_{novel} = \O$.
The \textit{base} set is used to pre-train a model and the \textit{novel} dataset is used to sample few-shot \underline{T}asks for evaluation.

In each \underline{T}ask, there are $N_c$ classes and each class has $N_s$ training samples, \ie, $N_c$-way $N_s$-shot, and the training samples are termed as support, which is used to differentiate from the training data in fully-supervised learning.
Also, the support may not be used to finetune the network. In prototype classification, the features of support videos are averaged for each class and the averaged features are used as prototype, \ie, the weights in the classifier, and the classification is done through nearest neighbor search.
Then, the test data within each \underline{T}ask is termed as query. Usually, during evaluation, each class have $15$ queries to be classified, \ie, $15 N_C$ queries in total.
During testing, for our approach, since we cannot aggregate the sentences into one sentence, we just calculate the distance between the query and all support videos, and use the distance averaged for each class is reference for classification.

Following the common setting~\citep{zhu2018compound}, we set $N_c=5$ and $N_s = \{1,5\}$, \ie, 5-way 1-shot and 5-way 5-shot. We then report the mean accuracy over 10,000 tasks for each task type. 
To note, as $\mathcal{C}_{base} \cap \mathcal{C}_{novel} = \O$, and no finetuning on support videos in \textit{novel} is applied, the recognition accuracy can be used to indicate the generalization ability of our approach on new classes.

For the dataset something-something v2 used in our experiment, and class typically asks the relationship between objects, such as something pushes something and the objects can vary from each other in different video instanes. As such, successful recognition on this dataset specifically requires the understanding of global temporal information.

\begin{table}[!htb]
    \centering
    \caption{Detailed results of performance (\%) on action recognition}
    \begin{tabular}{c|cc}
        \hlineB{3}
        Accuracy &  1-shot & 5-shot \\ \hline
        TempCLR w/ OTAM & 47.8 & 62.9 \\
        TempCLR w/ TRX & 45.1 & 67.7 \\
        \hlineB{3}
    \end{tabular}
    \label{tab:tempclr-fsac-supp}
\end{table}

For the results reported in few-shot action recognition, we use the model pre-trained under our self-supervision pre-training objective as the initialization and then use the loss proposed in OTAM or TRX for meta-training. The details of the results are reported in Table~\ref{tab:tempclr-fsac-supp}. We note that the initialization for all experiments and all approaches are the same for fair comparison. In this way, we highlight that our approach is orthogonal to the existing meta-training approaches and can significantly improve the few-shot adaptation efficiency. Meanwhile, the performance of TempCLR w/ OTAM on 5-shot is limited by the computational resources by the time of submission. We will provide more results when we access more GPU resources. 

\subsection{More Experimental Results}

\subsubsection{Measurement}

(\setup{Cap. Avg.}) still utilizes the unit-level comparison, \ie, each caption can be used to match the clip first. Then, for each video, the number of clips video retrieved by the captions in the set is used to indicate the similarity between the paragraph and the full video

(\setup{DTW}) Please refer to Sec.~\ref{sec:approach}. In particular, for the matching between units, DTW has a strong assumption that $\matchmat{1,1} = \matchmat{N_a,N_p} = 1$ is always met.

(\setup{OTAM}) stands for Ordered Temporal Alignment Module.  To avoid the boundary assumption in DTW, \citet{cao2020few} proposed OTAM. The basic idea is to pad the sequence with meaningless unit such as zero vector at the begining and end part. In this way, when dynamic programming is resued to calculate the matching, the first (last) units in the two sequences to be compared are not necessary matched. For detail about the implementation of OTAM, please refer to \citet{cao2020few}.

\subsubsection{Video retrieval on DiDeMo}

\revise{
We conduct video retrieval experiments on another dataset DiDeMo~\citep{anne2017localizing}, which contains 10,000 videos annotated with 40,000 sentences on Flicker videos. We follow VideoCLIP~\citep{xu2021videoclip} and conduct evaluation on 4021 test samples. 
}

\revise{
As summarized in Table~\ref{tab:retrieval-didemo-clip}, we first conduct experiments on the clip-caption retrieval setting. We follow VideoCLIP and summarized the performance in both zero-shot and supervised. Our \approach{} can consistently improve the performance than the VideoCLIP baseline and even outperform most supervised-based approaches. In particular, the performance at R@5 also outperform the best performance of ClipBERT.
}

\begin{table}[!ht]
    \centering
    \caption{Performance (\%) comparison on video retrieval on DiDeMo (clip-caption)}
    \begin{tabular}{lcc}
        \hlineB{3}
        Approach         & R@1  & R@5 \\ \hline
        \multicolumn{3}{c}{Supervised} \\ \hline
        S2VT~\citep{venugopalan2014translating} & 11.9 & 33.6 \\
        FSE~\citep{zhang2018cross} & 13.9 & 44.5 \\
        CE~\citep{liu2019use} & 16.1 & 41.1 \\
        ClipBERT~\citep{lei2021less} & {20.4} & {48.0} \\ \hline
        \multicolumn{3}{c}{Zero-shot transfer} \\ \hline
        VideoCLIP~\citep{xu2021videoclip} & 16.6 & 46.9 \\
        VideoCLIP$^{\dagger}$ & 16.4 & 47.1 \\
        \approach{} & \textbf{17.7} & \textbf{48.1} \\
        \hlineB{3}
    \end{tabular}
    \label{tab:retrieval-didemo-clip}
\end{table}

\revise{
Then, as compared in Table~\ref{tab:retrieval-didemo-full} \&~\ref{tab:retrieval-didemo-bg-full}, we conduct experiments on full-video retrieval (background removed). As the domain between HowTo100M can be distant from the domain of DiDeMo, the retrieval scores are generally lower than those on YouCookII, though the video in DiDeMo is shorter. 
In this case, we observe that exploring the temporal is very useful for the VideoCLIP baseline and our \approach{}. When the background is removed, Then, our approach \approach{} can consistently improve the performance by using DTW or OTAM as measure. In particular, using OTAM as measure can already achieve higher performance than using Cap. Avg. as measure based on unit-level comparison. At the same time, with Cap. Avg as measure, the retrieval performance is also improved comparing with VideoCLIP baseline.
Then, we also perform metric ensembling. For both VideoCLIP and TempCLR, employing metric ensembling can help improve the performance while our \approach{} is consistently better than the VideoCLIP baseline.
Finally, for full-video retrieval (background kept), for both VideoCLIP and \approach{}, using either DTW or OTAM as measure can improve the performance consistently than using Cap. Avg. as the measure. Again, as mentioned before, the domain gap between HowTo100M and DiDeMo may make the network difficult to generalize well, which then demonstrates the importance of using temporal context and perform sequence-level comparison directly. Then, TempCLR achieves the best performance when the metrics are ensembled. 
}

\begin{table}[!ht]
    \centering
    \caption{Performance (\%) comparison of full-video retrieval on DiDeMo (background removed)}
    \begin{tabular}{lccc}
        \hlineB{3}
        Measure         & R@1  & R@5  & R@10 \\ \hline
        \multicolumn{4}{c}{VideoCLIP} \\ \hline
        DTW             & 8.5 & 24.4 & 37.7 \\
        OTAM            & {8.9} & 26.6 & {38.6} \\
        DTW + Cap. Avg. & 10.3 & \textbf{28.2} & 38.6 \\
        OTAM + Cap. Avg.& 10.6 & 27.7 & 38.8 \\
        Cap. Avg.       & {8.9} & {27.0} & 37.9 \\ \hline
        \multicolumn{4}{c}{\approach{}} \\ \hline
        DTW             & 9.2 & 26.6 & 39.4 \\
        OTAM            & 10.4 & 25.6 & 39.9 \\
        DTW + Cap. Avg. & 9.9 & 26.3 & 38.4 \\
        OTAM + Cap. Avg.& \textbf{10.9} & 25.6 & \textbf{40.6} \\
        Cap. Avg.       & 9.4 & 26.3 & 36.4 \\
        \hlineB{3}
    \end{tabular}
    \label{tab:retrieval-didemo-full}
\end{table}

\begin{table}[!ht]
    \centering
    \caption{Performance (\%) comparison of full-video retrieval on DiDeMo (background kept)}
    \begin{tabular}{lccc}
        \hlineB{3}
        Measure         & R@1  & R@5  & R@10 \\ \hline
        \multicolumn{4}{c}{VideoCLIP} \\ \hline
        DTW             & 8.6 & 21.0 & 31.2 \\
        OTAM            & 8.1 & 22.1 & 31.7 \\
        DTW + Cap. Avg. & 9.0 & 22.6 & 32.4 \\
        OTAM + Cap. Avg.& 8.3 & 22.6 & 32.6 \\
        Cap. Avg.       & 7.1 & 21.0 & 31.9 \\ \hline
        \multicolumn{4}{c}{\approach{}} \\ \hline
        DTW             & 9.0 & 21.4 & 31.0 \\
        OTAM            & 9.3 & 21.4 & 31.4 \\
        DTW + Cap. Avg. & 9.0 & 21.7 & 31.2 \\
        OTAM + Cap. Avg.& \textbf{9.5} & \textbf{21.7} & \textbf{31.4} \\
        Cap. Avg.       & 8.1 & 21.7 & 32.8 \\
        \hlineB{3}
    \end{tabular}
    \label{tab:retrieval-didemo-bg-full}
\end{table}

\subsubsection{Full-Video retrieval}

As shown in Table~\ref{tab:retrieval-appendix}, we comprehensively study the performance of our approach under different metrics. 
Again, as VideoCLIP is not trained to explore the global temporal context, there is a large performance drop when DTW or OTAM is used to measure the distance. For video retrieval where the background is removed, the number of features in the paragraph part and the video part are the same where the features with the same index should be matched. As DTW has a strong assumption that $M(1,1) = M(N_a,N_p) = 1$, such prior constraint can benefit the matching.

However, for our approach \approach{}, when the model has been trained to compare the sequences directly, using either OTAM or DTW and always achieve good performance. In addition, Cap. Avg is unit-level based metric and can serve as a complementary component for video retrieval. As such, DTW/OTAM and can combined with Cap. Avg and the prediction by the combined metric can also achieve very high recall. 
Nevertheless, we believe the temporal alignment between two sequences still take the dominant role since the performance by the ensembled metric does not provide significant gain over the recall by using DTW/OTAM only. 

\begin{table}[h]
    \centering
    \caption{Ablation study on full-video retrieval (Background Removed)}
    \begin{tabular}{lccc}
        \hlineB{3}
        Measure         & R@1  & R@5  & R@10 \\ \hline
        \multicolumn{4}{c}{VideoCLIP} \\ \hline
        DTW             & 56.0 & 89.9 & 96.3 \\
        OTAM            & 52.8 & 89.2 & 95.0 \\
        DTW + Cap. Avg. & 85.8 & 97.7 & 99.1 \\
        Cap. Avg.       & 74.5 & 94.1 & 97.9 \\ \hline
        \multicolumn{4}{c}{\approach{}} \\ \hline
        DTW             & 83.5 & 97.2 & 99.3 \\
        OTAM            & 84.9 & 97.9 & 99.3 \\
        DTW + Cap. Avg. & 86.5 & 97.2 & 99.3 \\
        Cap. Avg.       & 74.5 & 94.6 & 97.0 \\
        \hlineB{3}
    \end{tabular}
    \label{tab:retrieval-appendix}
\end{table}

\revise{
Then, the performance of full-video retrieval (background kept) is summarized in Table.~\ref{tab:retrieval-bg-appendix}. As VideoCLIP has been well-trained on the full HowTo100M dataset and compare clip-caption pairs across videos, VideoCLIP can thus discriminate clips of background and can maintain high performance (\ie, 73.6) with Cap. Avg. as the measure. 
However, as the clips of background accounts for most part of the video and VideoCLIP is not trained to model the temporal context, the performance drop clearly when DTW or OTAM is used as measure. Furthermore, when we ensemble the DTW-related measure with Cap. Avg., the performance improvement is thus limited (\ie, from 73.6 to 74.8 by OTAM and 73.8 by DTW).}

\revise{
However, when we train the TempCLR on a very small dataset, the model can then capture the temporal context information effectively. We note that DTW has strong boundary assumption and thus the performance is limited. However, OTAM can avoid the assumption and can then improve R@1 of TempCLR from 70.4 to 72.2. In this way, we believe necessary modification on DTW-related measure can be done to further improve the training efficiency and we leave this for future work. In addition, we can achieve higher performance when the measures are combined. For example, by combining OTAM and Cap. Avg., the performance can be improved to 77.5, which further demonstrates that purely using Cap. Avg. as measure is not enough and it is necessary to explore the temporal context in feature embedding.}

\begin{table}[h]
    \centering
    \caption{Ablation study on full-video retrieval (Background Kept)}
    \begin{tabular}{lccc}
        \hlineB{3}
        Measure         & R@1  & R@5  & R@10 \\ \hline
        \multicolumn{4}{c}{VideoCLIP} \\ \hline
        DTW             & 55.7 & 93.1 & 98.9 \\
        OTAM            & 56.6 & 92.8 & 98.9 \\
        DTW + Cap. Avg. & 73.8 & 95.4 & 98.6 \\
        OTAM + Cap. Avg.& 74.5 & 95.1 & \textbf{99.1} \\
        Cap. Avg.       & 73.6 & 94.7 & 98.4 \\ \hline
        \multicolumn{4}{c}{\approach{}} \\ \hline
        DTW             & 70.4 & 93.8 & 97.9 \\
        OTAM            & 72.2 & 94.5 & 97.7 \\
        DTW + Cap. Avg. & 76.7 & 95.6 & 98.4 \\
        OTAM + Cap. Avg.& \textbf{77.5} & \textbf{95.6} & 98.6 \\
        Cap. Avg.       & 71.7 & 94.5 & 97.9 \\
        \hlineB{3}
    \end{tabular}
    \label{tab:retrieval-bg-appendix}
\end{table}

\subsubsection{Action step localization}
As VideoCLIP employs a Transformer architecture, we also study two different training strategies, \ie, updating all parameters and updating the layer norms only. 

(\setup{Zero-Shot}) The results in Table~\ref{tab:Localization-appendix} (Left) study the performance by training on HW100M using either OTAM or DTW as to measure sequence distance in \approach{}. It is observed that finetuning all parameters can improve the zero-shot adaptation performance most and using either OTAM or DTW achieves similar performance. However, if only the norm layer is updated, the trianing efficiency by using OTAM as sequence distance measurement can be slightly less effective.

(\setup{Supervised}) For the finetuning on CrossTask, we always update the whole network for fair comparison. However, the pre-trained model by our \approach{} is obtained by updating either all parameters or norm layer only during the pre-traning stage. The two pre-trained models uses DTW as measurement in $\mathcal{L}_{seq}$. In the Table~\ref{tab:Localization-appendix} (Right), each column indicates how the finetuning loss is set where w/o $\mathcal{L}_{seq}$ indicates the vanilla contrastive-learning based approach between captions and clips which does not consider the global temporal context.

Then, we can find that 1) using DTW during finetuning is better than using OTAM and 2) updating the layer norm only can stably improve the performance with small variance. When DTW is used and all parameters are finetuned, the model can be well-learned during the pre-training stage and thus achieves good performance in the \setup{Supervised} setup. 

\begin{table}[h]
    \caption{Ablation study on action step localization under \setup{Zero-Shot} (Left) and \setup{Supervised} (Right).}
    \parbox{.43\textwidth}
    {
        \centering
        \resizebox{.43\textwidth}{!}
        {%
            \renewcommand{\arraystretch}{1.13}
            \begin{tabular}{l|cc}
                \hlineB{3}
                (\setup{Zero-Shot})  & OTAM & DTW  \\ \hline
                \approach{} + All Params. & 36.9 & 36.9 \\
                \approach{} + Norm Only   & 36.1 & 36.8 \\
                \hlineB{3}
                \multicolumn{3}{l}{The performance of VideoCLIP is 33.9}
            \end{tabular}
        }
    }
    \hfill
    \parbox{.55\textwidth}
    {
        \centering
        \resizebox{.55\textwidth}{!}
        {%
            \renewcommand{\arraystretch}{1.13}
            \begin{tabular}{l|ccc}
                \hlineB{3}
                (\setup{Supervised})  & w/o $\mathcal{L}_{seq}$ & OTAM & DTW  \\ \hline
                VideoCLIP              & 47.3      & 51.7 & 52.5 \\
                \approach{} + All Params. & 49.3        & 50   & 52.3 \\
                \approach{} + Norm Only   & 49.0      & 51.1 & 51.6 \\
                \hlineB{3}
            \end{tabular}
        }
    }\label{tab:Localization-appendix}
\end{table}

Then, we provide the full table of \setup{Zero-Shot} in Table~\ref{tab:Localization-appendix-full}.
As mentioned in the main paper, VideoCLIP~\citep{xu2021videoclip} does not provide clear performance gain over HT100M and under-performs many other SOTA approach in zero-shot evaluation. Since we only have 90k training data and is trained from VideoCLIP, the model by our approach is hard to be well-learned for the zero-shot localization evaluation. However, our approach can still achieve 9\% gain and improve the recall from 33.9 to 36.9.
Nevertheless, as summarized in Table~\ref{tab:finetune-full}, by finetuning from 50 to 100 training data on CrossTask, we can effectively improve the performance.

\begin{table}[h]
    \centering
    \caption{Full table for Action Step Localization (\setup{zero-shot})}
    \begin{tabular}{l|cc}
        \hlineB{3}
        Approach (\setup{Zero-Shot}) & Backbone & Recall \\
        \hline
        HT100M~\citep{miech2019howto100m}  & R152 + RX101 & 33.6 \\
        MIL-NCE~\citep{miech2020end}     & S3D-G        & 40.5 \\
        MIL-NCE~\citep{miech2020end}     & I3D-G        & 36.4 \\
        ActBERT~\citep{zhu2020actbert}     &   R101 + Res3D           & 37.1 \\
        ActBERT~\citep{zhu2020actbert}     &   + Faster-RCNN           & 41.4 \\
        UniVL~\citep{luo2020univl}       &   S3D-G    & 42.0 \\
        MCN~\citep{chen2021multimodal}         & R152 + RX101 & 35.1 \\ 
        MMFT~\citep{shvetsova2022everything}        & R152 + RX101  & 39.3 \\
        MMFT~\citep{shvetsova2022everything}        &   S3D-G     & 41.1 \\
        \hline
        VideoCLIP~\citep{xu2021videoclip}   &      S3D-G        & 33.9 \\
        \approach{} (Ours)        &   S3D-G           & 36.9 ($\uparrow \mathbf{3.0}$) \\
        \hlineB{3}
    \end{tabular}
    \label{tab:Localization-appendix-full}
\end{table}

\begin{table}[h]
    \centering
    \caption{Few-shot fine-tuning for action step localization}
    \begin{tabular}{c|cc}
        \hlineB{3}
        Approach & 50(10\%) & 100(20\%) \\
        \hline
        VideoCLIP~\citep{xu2021videoclip} &  40.1 & 41.1 \\
        \approach{} (Ours) &  40.2	& 42.8 \\
        \hlineB{3}
    \end{tabular}
    \label{tab:finetune-full}
\end{table}

\subsubsection{Discussion on Negative Size}

\revise{
As summarized in Table~\ref{tab:negative-size}, we adjust the size of negative samples from 16 to 64 and use the zero-shot performance on action step localization and full-video retrieval for comparison. We use Recall (\%) as metric for action step localization and R@1 (\%) for full-video retrieval evaluation.
By increasing the negative size from 16 to 32, we can see the performance can be improved slightly. However, when we keep increasing the size to 64, there is no significant gain. To note, during pre-training, as we directly use the data sampling strategy in VideoCLIP, the number of segments existing in each video is only 6 on average. 
To reach reasonably high performance but still train the network efficiently, we set the size of negative samples as 32.
}

\begin{table}[]
    \centering
    \caption{Ablation study of Negative Size in Zero-shot transfer learning (\%)}
    \begin{tabular}{c|ccc}
    \hlineB{3}
        Task $\backslash$ Negative Size & 16 & 32 & 64 \\ \hline
        action step localization (Recall)& 36.5 & 36.9 & 37.1 \\
        full-video retrieval (R@1) & 82.9 & 83.5 & 83.5 \\
    \hlineB{3}
    \end{tabular}
    \label{tab:negative-size}
\end{table}

\subsubsection{Discussion on Unit Matching in Sequence Alignment}\label{sec:gradient}

From the results in Table~\ref{tab:retrieval-appendix} and Table~\ref{tab:ablation}(a), video clips relies mainly replies on the unit-level comparison for video retrieval (hige recall by Cap. Avg.). Even though the percentage of corrected matched caption-clips between the paragraph and video in the pair reaches 80\%, the recall is still low when DTW/OTAM is used. This is because the set of sentences features can also have high alignment score with (\ie, distracted by) clip features from other videos, which further demonstrate that the success of VideoCLIP is mainly based on unit-level comparison. However, our approach \approach{} improves both recall and the percentage of correctly matched caption-clip pairs, which indicates that the model has indeed been trained to model the global temporal correlation. 

%Also, when the annotation is not available, according to the visualization from OTAM, the matched videos are also closed to each other given the action background (the visualization in OTAM~\cite{cao2020few}). As such, using OTAM or DTW can provide a weakly supervised signal for the case where no or noisy label is provided.
%  AlignNet~\cite{wang2020alignnet} has been proposed to provide dense alignment between continuous video and audio signals. It focus on the multi-scale (for diverse distortion and speed change) and use layers of different level to wrap and integrade the bottom layer with current layer, which is then for cross-modal analysis. Our approach is for sparse alignment and matching. but orthognal w.r.t. each other.

\subsection{Comparison of Optimization Gradient between TempCLR and VideoCLIP}

\revise{
In this section, to explain the difference of training objective in VideoCLIP and TempCLR, we compare the gradient by obtained by different training strategies, temporally overlapped strategy by VideoCLIP~\citep{xu2021videoclip} and the DTW-based sequence-level comparison by \approach{}. We take a toy example where two sequences $M=[m_1,m_2]$ and $N=[n_1,n_2]$ should be aligned and the ground-truth of matched pairs are $(m_1,n_1)$ and $(m_2,n_2)$. Without loss of generality, we set $M$ as anchor and $N$ as query. Then, the training losses are,}

(for VideoCLIP)
\[
\mathcal{L}_{VideoCLIP} = \mathcal{L}_1 + \mathcal{L}_2 = -\log\frac{e^{m_1 \cdot n_1^T}}{e^{m_1 \cdot n_1^T} + e^{m_1 \cdot n_2^T}} - \log\frac{e^{m_2 \cdot n_2^T}}{e^{m_2 \cdot n_2^T} + e^{m_2 \cdot n_1^T}}
\]
where $\mathcal{L}_1$ and $\mathcal{L}_2$ are the losses of positive pairs for anchor $m_1$ and $m_2$ in $M$ respectively,

and (for TempCLR) 
\[
\mathcal{L}_{TempCLR} = -\log\frac{e^{m_1 \cdot n_1^T + m_2 \cdot n_2^T}}{e^{m_1 \cdot n_1^T + m_2 \cdot n_2^T} + e^{m_1 \cdot n_2^T + m_2 \cdot n_1^T}}
\]
where $\mathcal{L}_{TempCLR}$ is the loss for $M$.

Then, we take the positive pair $(m_1,n_1)$ as one example and derive the gradients w.r.t. $m_1 \cdot n_1^T$,

(For VideoCLIP)
\[
\frac{\partial \mathcal{L}_{VideoCLIP}}{ \partial(m_1 \cdot n_1^T)} = \frac{\partial \mathcal{L}_1}{ \partial(m_1 \cdot n_1^T)} = -e^{m_1 \cdot n_1^T}\frac{e^{m_1 \cdot n_2^T}}{e^{m_1 \cdot n_1^T}(e^{m_1 \cdot n_1^T}+e^{m_1 \cdot n_2^T})} = \frac{-e^{m_1 \cdot n_2^T}}{e^{m_1 \cdot n_1^T}+e^{m_1 \cdot n_2^T}}
\]
and (For TempCLR)
\[
\frac{\partial \mathcal{L}_{TempCLR}}{ \partial(m_1 \cdot n_1^T)} = -e^{m_1 \cdot n_1^T + m_2 \cdot n_2^T}\frac{e^{m_1 \cdot n_2^T + m_2 \cdot n_1^T}}{e^{m_1 \cdot n_1^T + m_2 \cdot n_2^T}(e^{m_1 \cdot n_1^T + m_2 \cdot n_2^T}+e^{m_1 \cdot n_2^T + m_2 \cdot n_1^T})}
\]
which can be further simplified as
\[
\frac{\partial \mathcal{L}_{TempCLR}}{ \partial(m_1 \cdot n_1^T)} = \frac{-e^{m_1 \cdot n_2^T}}{e^{m_1 \cdot n_1^T}e^{m_2 \cdot n_2^T - m_2 \cdot n_1^T}+e^{m_1 \cdot n_2^T}}
\]

As such, for our \approach{}, we can observe that the optimization of the positive pair $m_1 \cdot n_1^T$ clearly considers the difference when $m_2$ is matched with each element in $N$, \ie, $e^{m_2 \cdot n_2^T - m_2 \cdot n_1^T}$. In other words, the optimization of $m_1 \cdot n_1^T$ also depends on the correlation between $m_2 \cdot n_2^T$ and $m_2 \cdot n_1^T$, which is the temporal context.
\[
\frac{\partial \mathcal{L}_{VideoCLIP}}{ \partial(m_1 \cdot n_1^T)}/\frac{\partial \mathcal{L}_{TempCLR}}{ \partial(m_1 \cdot n_1^T)} = \frac{(e^{m_1 \cdot n_1^T}+e^{m_1 \cdot n_2^T})^{-1}}{(e^{m_1 \cdot n_1^T}e^{m_2 \cdot n_2^T - m_2 \cdot n_1^T}+e^{m_1 \cdot n_2^T})^{-1}}
\]

We then take $m_1 \cdot n_2^T$ as one example and compare the optimization of negative pairs. 
(For VideoCLIP)
\[
\frac{\partial \mathcal{L}_{VideoCLIP}}{ \partial(m_1 \cdot n_2^T)} = \frac{\partial \mathcal{L}_1}{ \partial(m_1 \cdot n_2^T)} = \frac{e^{m_1 \cdot n_2^T}}{e^{m_1 \cdot n_1^T}+e^{m_1 \cdot n_2^T}}
\]
and (For TempCLR)
\[
\frac{\partial \mathcal{L}_{TempCLR}}{ \partial(m_1 \cdot n_2^T)} = \frac{e^{m_1 \cdot n_2^T + m_2 \cdot n_1^T}}{e^{m_1 \cdot n_1^T + m_2 \cdot n_2^T} + e^{m_1 \cdot n_2^T + m_2 \cdot n_1^T}}
\]

Similarly, we can also observe that optimization of negative pairs only considers unit-level comparison $m_1 \cdot n_1^T$ and $m_2 \cdot n_2^T$ in VideoCLIP while ours also takes temporal context, \ie, the matching difference $m_2 \cdot n_2^T - m_2 \cdot n_1^T$ into consideration. 
\[
\frac{\partial \mathcal{L}_{VideoCLIP}}{ \partial(m_1 \cdot n_2^T)}/\frac{\partial \mathcal{L}_{TempCLR}}{ \partial(m_1 \cdot n_2^T)} = \frac{(e^{m_1 \cdot n_1^T - m_1 \cdot n_2^T } + 1)^{-1}}{(e^{m_1 \cdot n_1^T - m_1 \cdot n_2^T + m_2 \cdot n_2^T - m_2 \cdot n_1^T} + 1)^{-1}}
\]

\subsection{More Visualization \& Figure Explanation}

\begin{figure}[!htb]
    \centering
    \includegraphics[width=\textwidth]{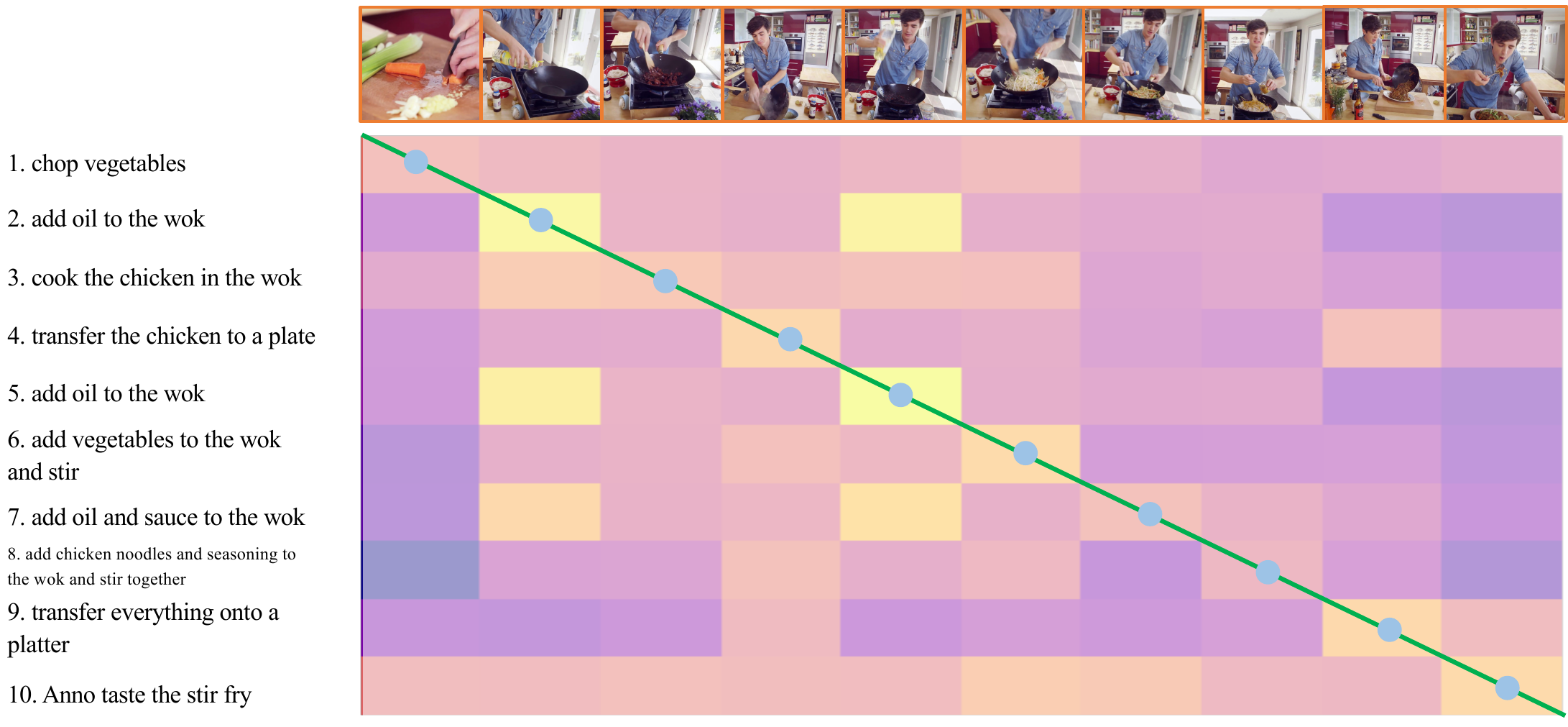}
    \caption{Full version of visualization in the main paper.}
    \label{fig:visualization-1}
\end{figure}

\begin{figure}[!htb]
    \centering
    \includegraphics[width=\textwidth]{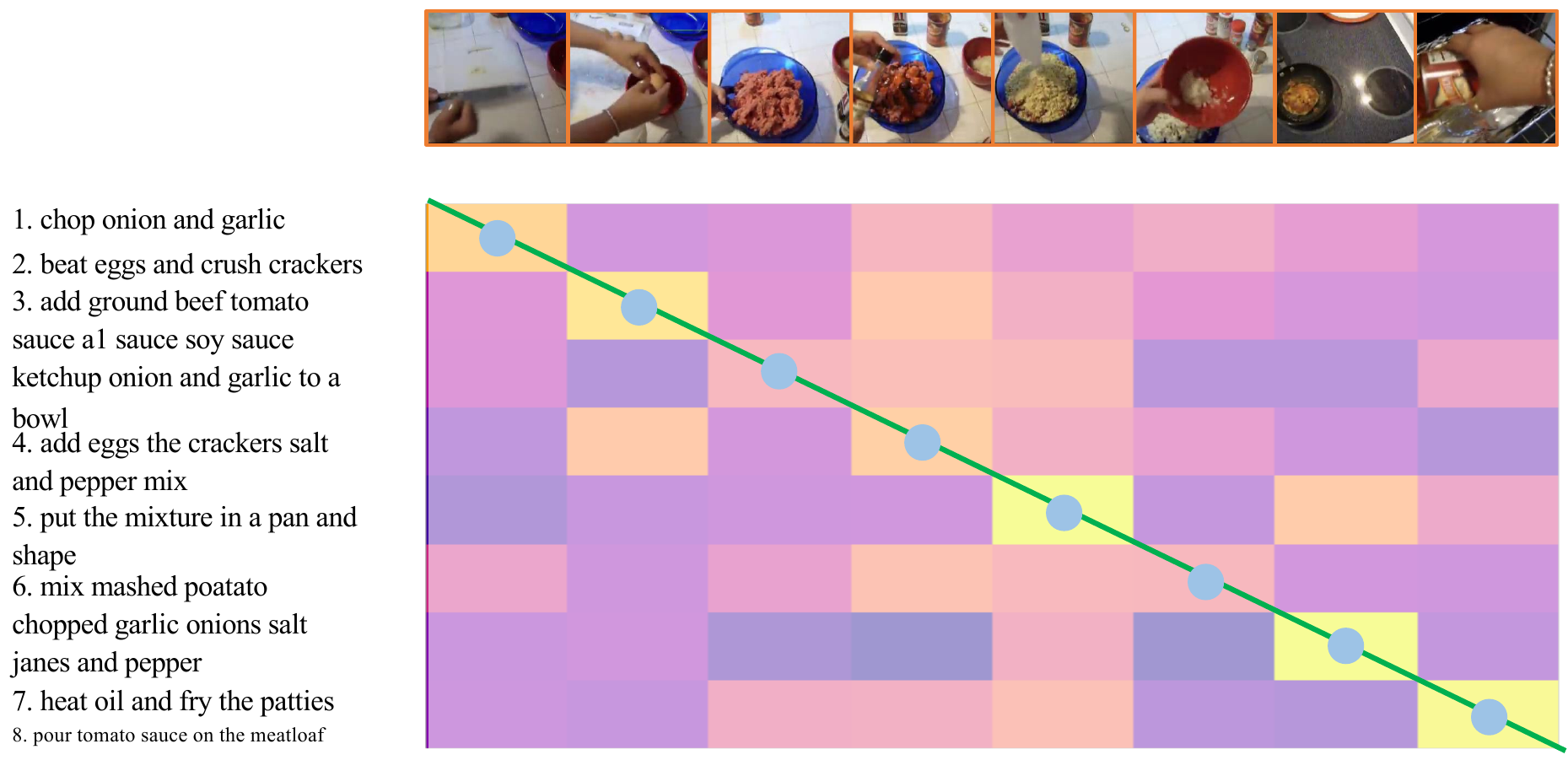}
    \caption{Additional visualizaton examples.}
    \label{fig:visualization-2}
\end{figure}

\begin{figure}[!htb]
    \centering
    \includegraphics[width=\textwidth]{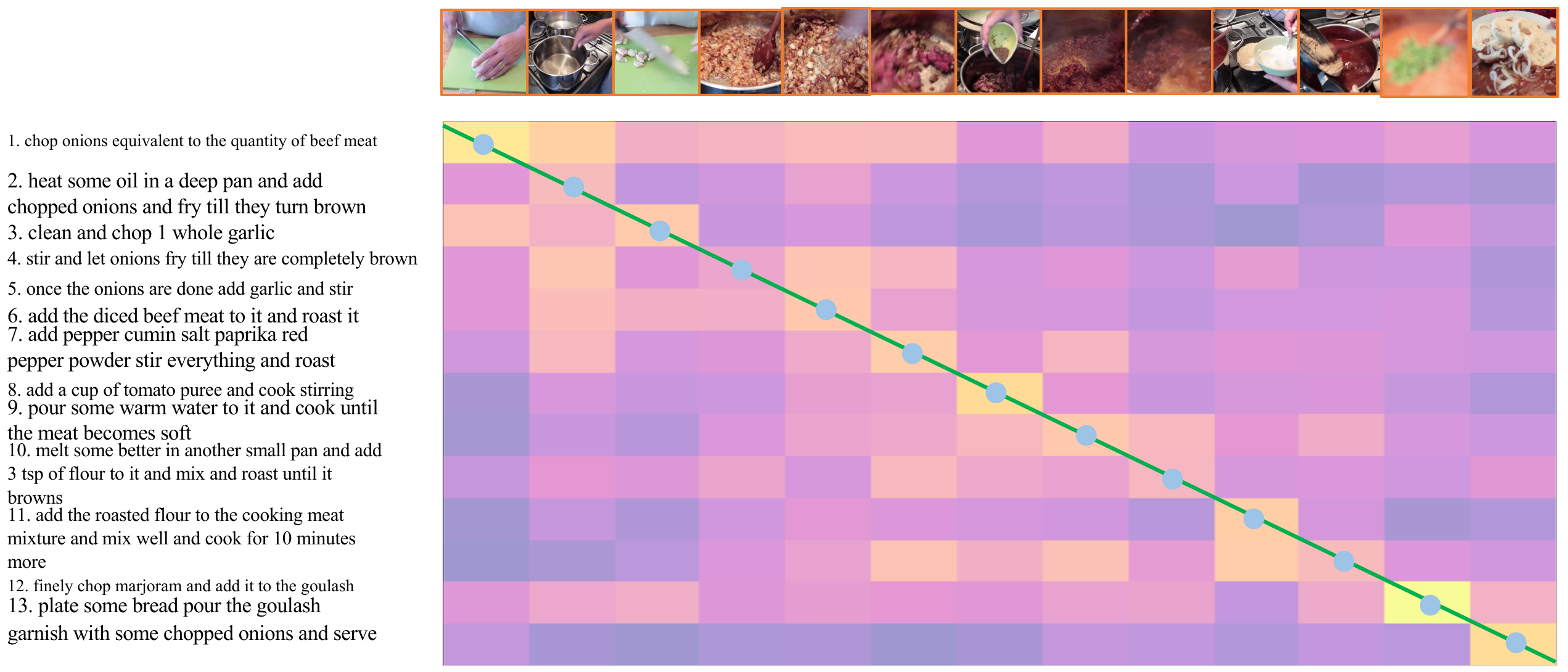}
    \caption{Additional visualizaton examples.}
    \label{fig:visualization-3}
\end{figure}

\begin{figure}[!htb]
    \centering
    \includegraphics[width=\textwidth]{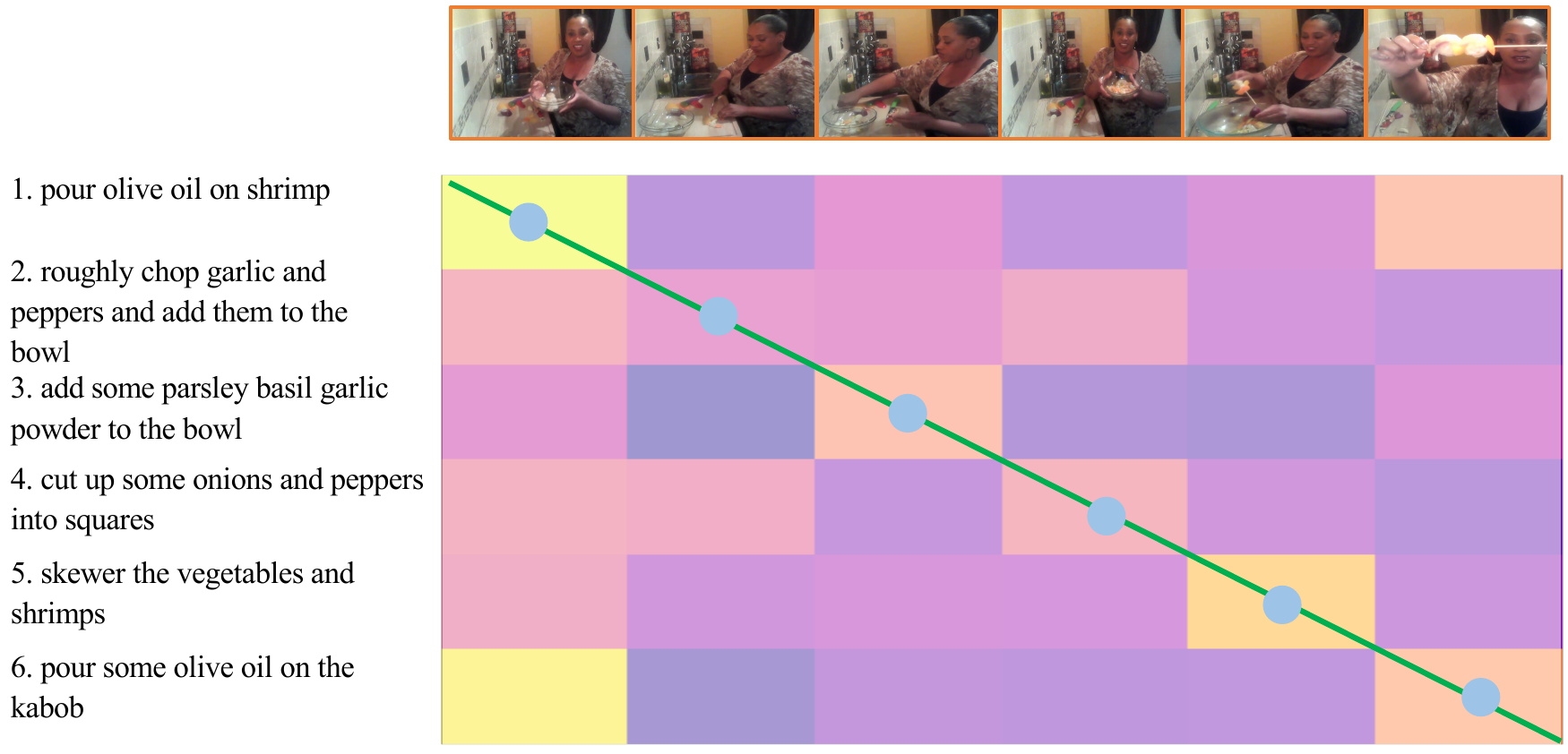}
    \caption{Additional visualizaton examples.}
    \label{fig:visualization-4}
\end{figure}

We first provide the full version of visualization shown in Fig.~\ref{fig:visualization-main} in Fig.~\ref{fig:visualization-1}. Then, we provide three more visualization results in Fig.~\ref{fig:visualization-2} \& \ref{fig:visualization-3} \& \ref{fig:visualization-4}. The visualization demonstrates that our approach is capable to correctly match the clip-caption pairs with high confidence and also achieve sequence alignment for full-video retrieval. 

Meanwhile, in our Concept Figure~\ref{fig:concept}, as a comparison between sequence-level distance and unit-level distance, (b) has similar color pattern to (a) since the color of chicken is dark, which results in difficult and potential mismatching under unit-level comparison.

\begin{figure}[!htb]
    \centering
    \includegraphics[width=\textwidth]{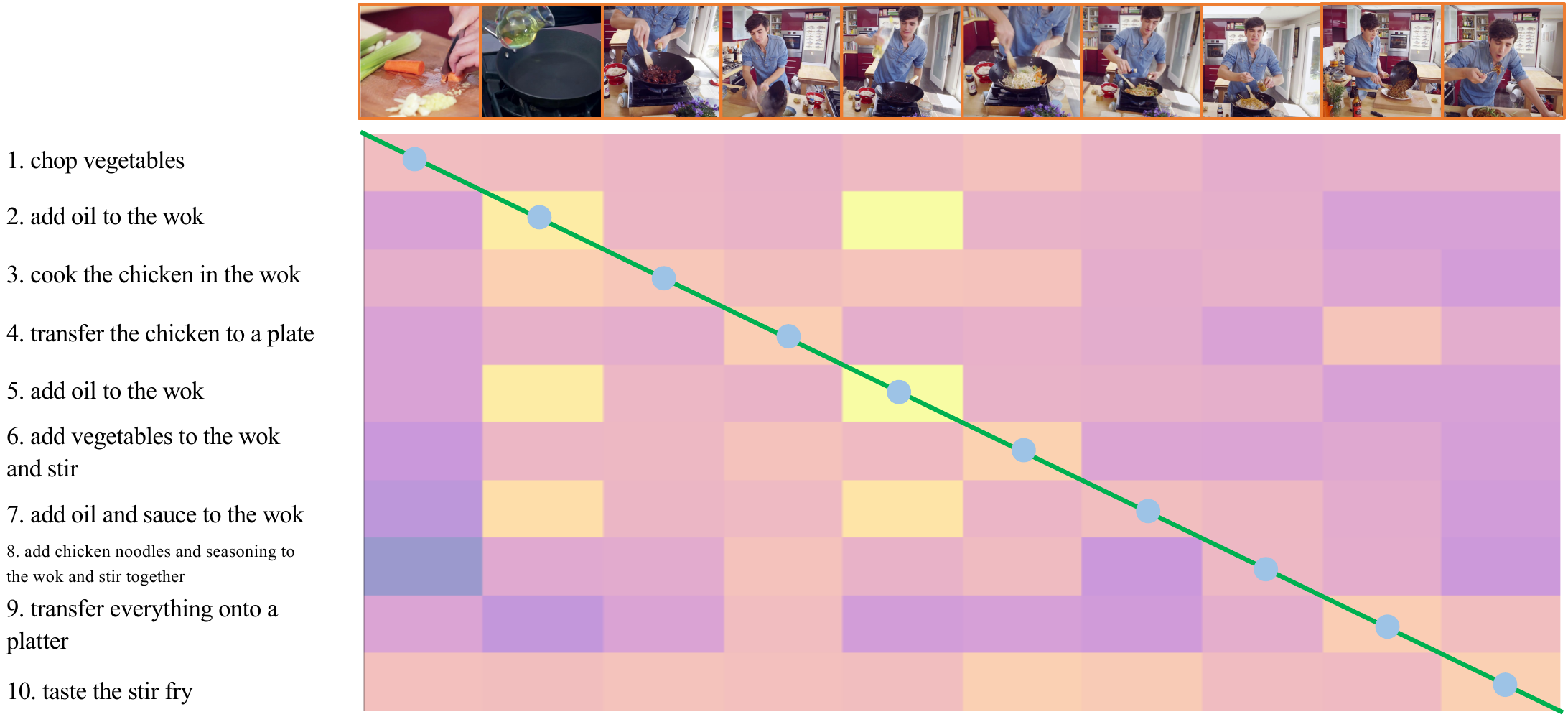}
    \caption{Visualization on scene change.}
    \label{fig:visualization-scene}
\end{figure}

\revise{
Then, we examine the robustness of our \approach{} towards scene change. As shown in Fig.~\ref{fig:visualization-scene}, we consider the case where the change of scene/context does not break the temporal succession. Given a full video which consists of 10 captions and each caption is paired with a segment in the full video. Then, we replace the video segment corresponding to the second step ``add oil to the wok'' as a segment which is also describing ``add oil to the wok'' but from another video. As we feed the whole video into the network jointly, the interaction between the clips of the replaced segment with others may adjust the extracted features, comparing with Fig.~\ref{fig:visualization-1}, the similarity between the captions and the original video clips are changed slightly. However, with DTW as measure, as the steps in the videos are still successive, we can still align the captions with videos correctly.}

\begin{figure}[!htb]
    \centering
    \includegraphics[width=\textwidth]{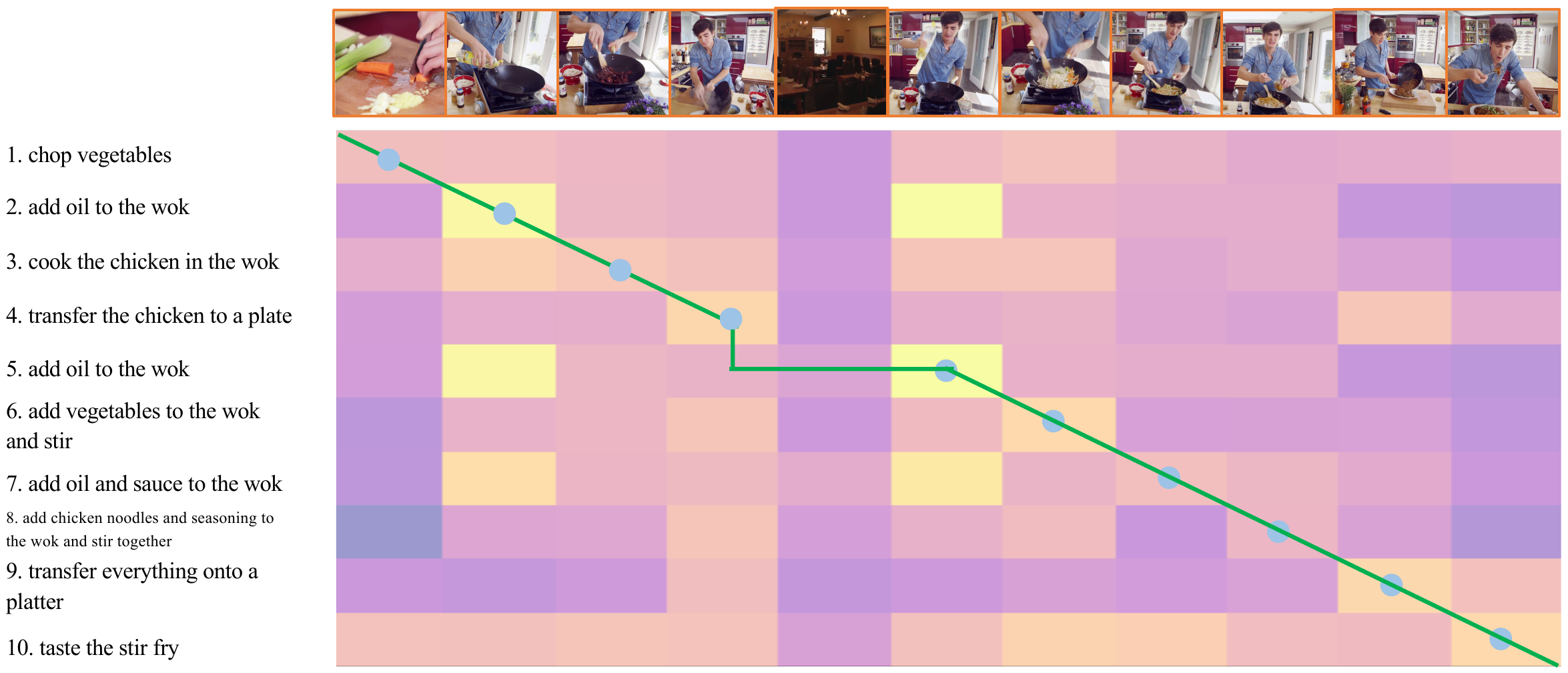}
    \caption{Visualization on background change.}
    \label{fig:visualization-background}
\end{figure}

\revise{
Next, we examine the robustness towards background (\ie, clips without clear semantic meaning), as shown in Fig.~\ref{fig:visualization-background}, as inject a segment of background between the 4th and 5th steps in the original video. Then, as the features of all captions are distant from the background, it does not impact the optimal alignment derived in DTW. As such, our \approach{} is still robust to the background.}

\end{document}